# Adapting Large Language Models via Multi-Task Learning for Mental Health Problem, Emotion, and Sentiment Detection

Amir Rafiq Bhat[1]
amir@stu.khas.edu.tr
0009-0006-6807-7495

Rahim Dehkharghani[1]
rahim.dehkharghani@khas.edu.tr
0000-0002-9619-8247

Zeki ERDEM[1,2]
zeki.erdem@khas.edu.tr
0000-0002-1441-076X

[1] Computer Engineering Department, Faculty of Engineering and Natural Sciences, Kadir Has University, Istanbul, Turkey
[2] Computer Science Department, College of Innovation and Technology, University of Michigan-Flint, Flint, MI, USA

***Abstract***— Mental health problem detection, emotion detection, and sentiment analysis are closely related tasks in Natural Language Processing (NLP). Recent advances in Large Language Models (LLMs) and multi-task learning (MTL) have shown strong potential for improving performance on such interconnected tasks. This paper investigates sequential multi-task learning using Parameter-Efficient Fine-Tuning (PEFT) with LoRA and 4-bit quantization on the Llama 3.1 8B Instruct model for Mental Health Problem (MHP) detection, multi-label emotion detection, and sentiment analysis. We provide a systematic empirical analysis of sequential multi-task learning with PEFT for mental health NLP and investigate how task ordering affects knowledge transfer and catastrophic forgetting across related affective computing tasks. In particular, we demonstrate that sequential performance trade-offs are highly sensitive to task order and yield significant performance differences across tasks. We evaluate two sequential fine-tuning orders: (1) Mental Health Problem → Emotion → Sentiment (MHP-first) and (2) Emotion → Mental Health Problem → Sentiment (Emo-first), using the extended SWMH dataset containing 4,243 English social media posts. Experimental results show that sequential PEFT significantly improves performance over the base LLM and reveals strong task-order dependencies. The MHP-first sequence achieved the best MHP detection performance with a Micro-F1 score of 0.7624, whereas the Emo-first sequence produced better Emotion (0.6052 F1) and Sentiment (0.9459 F1) results. The findings demonstrate that task ordering critically influences knowledge transfer, task interference, and overall performance in sequential MTL. The study highlights the effectiveness of PEFT-based LLM adaptation for mental health NLP while emphasizing the importance of carefully designing task sequences for related NLP tasks. In line with the focus on high-performance computing (HPC), adapting massive 8-billion-parameter foundation models imposes substantial memory and computational demands. We demonstrate that integrating sequential PEFT with 4-bit quantization reduces the number of trainable parameters and memory requirements associated with LLM adaptation, making the approach more suitable for GPU-accelerated and resource-constrained computing environments. Furthermore, to empirically validate the necessity of our sequential task ordering, we implement and evaluate a simultaneous Joint Multi-Task Learning (Joint-MTL) baseline, demonstrating the task interference and generative format collapse that occurs when LLMs are overloaded with complex, overlapping psychiatric paradigms.

Keywords — Large Language Models (LLMs), Multi-Task Learning (MTL), Mental Health Problem Detection, Emotion Detection, Sentiment Analysis

# 1. Introduction

The increasing global prevalence of mental health conditions presents one of the most significant public health challenges of the 21st century. According to the World Health Organization (WHO), approximately 970 million people (roughly 1 in 8) worldwide live with mental disorders [1]. However, many individuals do not seek help due to financial costs, accessibility issues, social stigma, or personal discomfort. This creates a pressing need for accessible, private, and scalable tools to support early detection.

Natural Language Processing (NLP) offers transformative potential by analyzing textual data from diverse sources such as social media and online forums to identify linguistic patterns associated with mental distress [19, 24]. Among these tools, Large Language Models (LLMs) such as the Llama series represent the current state of the art [2]. Trained on vast text corpora, they demonstrate a remarkable proficiency in understanding language. However, the power of LLMs comes with a challenge: they are inherently generalists. They are not specialized for the nuanced, sensitive, and often ambiguous language used to express mental health struggles. To perform optimally, they require adaptation. While full fine-tuning these multi-billion-parameter models is computationally prohibitive for most users, Parameter-Efficient Fine-Tuning (PEFT) techniques such as Low-Rank Adaptation (LoRA) [3] offer an efficient solution by updating only a small fraction of model parameters.

Furthermore, mental health is a multifaceted issue, involving a complex interplay of conditions, emotions, and sentiments. This makes it an ideal candidate for Multi-Task Learning (MTL), a paradigm where a model learns multiple related tasks to improve generalization.

Although prior studies have explored joint MTL for mental health using custom neural architectures [4], they generally overlook task-ordering effects, sequential adaptation dynamics, and PEFT strategies for LLMs. Unlike prior studies that primarily focus on joint MTL or single-task adaptation, this study investigates the sequential adaptation of a general LLM using PEFT. To the best of our knowledge, this paper is among the first to systematically characterize task-order-dependent transfer behavior in sequential PEFT adaptation of large language models for mental health NLP. Specifically, we examine these effects when adapting large language models to related tasks such as mental health problem detection, emotion detection, and sentiment analysis.

The motivation of the proposed methodology is illustrated in Figure 1. The figure highlights the progression from a general-purpose LLM to sequential task adaptation, emphasizing how the order of related tasks can influence knowledge transfer, forgetting, and overall task performance.

The main question this paper addresses is how the sequential exposure to different tasks influences model performance. We specifically examine the dynamics of knowledge transfer, the risk of task interference, and the critical impact of the fine-tuning sequence. This study focuses on a primary task, mental health problem detection, and two auxiliary tasks: multi-label emotion detection and binary sentiment analysis. The findings show that task ordering substantially affects the transfer behavior and the performance trade-offs in sequential PEFT adaptation for mental health NLP. Our contributions can be summarized as follows.

- Providing a systematic empirical analysis of sequential multi-task learning using parameter-efficient fine-tuning (PEFT) for mental health detection.

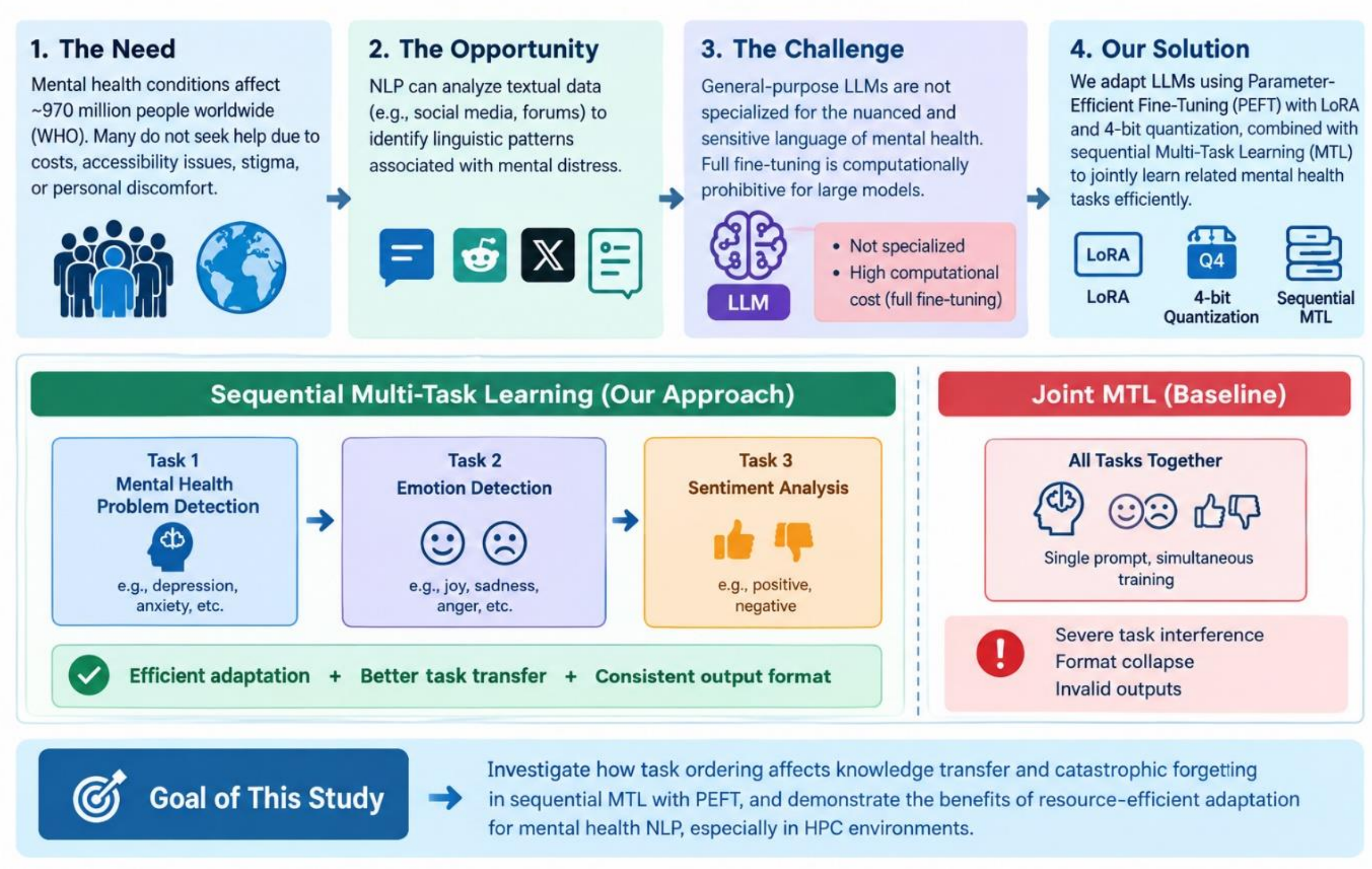

Figure 1: Motivation and overview of the proposed methodology.

- Quantifying the critical performance trade-offs inherent in task ordering, demonstrating that the optimal fine-tuning curriculum is highly task-dependent.
- Conducting a rigorous comparative analysis against a simultaneous Joint Multi-Task Learning (Joint-MTL) baseline, empirically demonstrating that the evaluated simultaneous multi-tasking configuration induces severe task interference and generative format collapse, highlighting the advantages of our proposed sequential approach.
- Analyzing the effects of task ordering on knowledge transfer across related affective computing tasks.
- Investigating the computational relevance of PEFT-based sequential LLM adaptation by reducing the number of trainable parameters and the memory footprint associated with adapting an 8-billion-parameter model.

The remainder of this paper is organized as follows. Section 2 reviews the related studies. Section 3 presents the methodology. Section 4 describes the experimental setup and results. Section 5 discusses the findings, and Section 6 addresses the ethical considerations of this study, and Section 7 concludes the paper.

# 2. Background and related work

This section reviews prior work on multi-task learning, PEFT-based LLM adaptation, and NLP applications in Mental health problem detection, and the specific considerations of sequential learning. The following subsections report related work to each field of study.

### *2.1. Multi-Task Learning Paradigms*

Multi-Task Learning (MTL) aims to improve model generalization and data efficiency by learning multiple related tasks simultaneously or in sequence, leveraging shared representations [5] [6]. Joint MTL typically involves hard parameter sharing (shared lower layers, task-specific heads) or soft parameter sharing (regularizing parameters of separate task models to be similar) [7]. Sequential MTL, the focus of this work, involves fine-tuning a model on a sequence of tasks, where knowledge from earlier tasks can potentially provide an inductive bias for subsequent ones [8] [9]. While offering a simpler fine-tuning process and the potential for curriculum-like benefits, sequential MTL faces the challenge of catastrophic forgetting and task interference, where learning new tasks can degrade performance on previously learned ones. [10]

Recent studies have focused intensely on this issue in the context of modern LLMs. For instance, Haque et al. [13] provide a comparative analysis of catastrophic forgetting across various LLMs in continual fine-tuning scenarios. To counteract this, novel technical solutions are being developed. Song et al. [14] proposed a hierarchical regularization framework that computes element-wise parameter importance to protect general knowledge during fine-tuning. Similarly, Zhang et al. [15] introduced Task-Adaptive Low-Rank Representation (TA-LoRA), an advanced MTL method that employs a fast-slow weights mechanism to more effectively separate shared and task-specific knowledge, thereby mitigating knowledge degradation.

### *2.2. LLM Adaptation with PEFT*

The advent of LLMs such as Llama 3.1 [2] has provided powerful foundational models. However, their large scale makes full fine-tuning prohibitively expensive. PEFT methods address this by updating only a small subset of parameters. Low-Rank Adaptation (LoRA) [3] freezes pre-trained weights and injects small, trainable low-rank matrices into specific layers. Quantized Low-Rank Adaptation (QloRA) [16] further enhances efficiency by combining LoRA with 4-bit quantization of the base model, making fine-tuning of very large models more accessible, as used in this study. The field of PEFT is rapidly evolving, as shown in prior work such as Wang et al. [17], which provides a detailed taxonomy of various approaches, including additive, selective, and reparametrized methods. Building on these foundations, the solutions proposed by both Song et al.[14] and Zhang et al. [15] integrate their forgetting-mitigation strategies directly with LoRA, demonstrating that PEFT is a critical component for developing robust sequential learning systems. Other PEFT techniques include adapter tuning (Houlsby et al.) [18], prompt-based methods like prefix-tuning [19] and prompt tuning [20]. While recent 2025 studies either propose novel algorithmic modifications to PEFT architectures to mitigate forgetting [14, 15] or conduct broad comparative analyses across general NLP benchmarks [13], our work is a highly focused empirical application study. We aim to characterize the baseline sequential transfer behavior and task-interference inherent to the standard QLoRA specifically within the interconnected affective computing domain. Furthermore, recent advances in transferable representation learning and optimization-aware fine-tuning [34] [35] [36] [37] provide valuable context for efficient parameter adaptation in large-scale models.

### *2.3. Natural Language Processing and Multi-Task Learning in Mental Health*

NLP techniques are increasingly applied to textual reviews in social media to analyse mental health insights, for detection of mental health problems such as depression, anxiety, and suicidality, as well as associated emotions and sentiment [21] [22] [23]. The choice of adaptation strategy is a critical research question. Kermani et al. [24] systematically compared fine-tuning, Retrieval-Augmented Generation (RAG), and prompt engineering for mental health text analysis, finding that while full fine-tuning offers the highest accuracy, its resource demands motivate the use of more computationally efficient methods such as PEFT techniques employed in our study. The SWMH dataset [26] is a popular benchmark for detecting mental health problems in the literature. Armah and Dehkharghani [4] extended a 5,000-instance subset of SWMH with emotion and sentiment labels [26] and proposed custom joint MTL architectures (Conv1D/BiLSTM-based) for simultaneously detecting mental health problems, emotions, and sentiment. Their work provides the dataset used in our study. Other works have also explored domain-specific LLMs for mental health, such as MentalBERT [27] and MentaLLaMA [28], often employing fine-tuning or MTL.

Recent studies have also validated joint multi-task learning approaches for simultaneously classifying mental health disorders, emotions, and sentiment. Pathak et al. [29] developed a Core Fusion Network (CFN) to jointly classify mental health disorders, emotions, and sentiment. Their results demonstrated that a tri-task setting leveraging the relationships between affective states and mental health conditions outperformed bi-task and uni-task approaches. Our work extends this line of inquiry by focusing on sequential adaptation, a direction that aligns with emerging research into the temporal dynamics of mental health. As another example, Ali et al. [30] proposed a longitudinal framework that treats multiple clinical interactions over time as a continuous conversation, highlighting the importance of tracking changes in patient states. While their work is multimodal, its focus on temporal progression aligns conceptually with the sequential nature of our study.

The inherent relatedness of mental state, emotional expression, and sentiment, as described by Akhtar et al. [31], makes this domain well-suited for MTL. Despite recent advances, important gaps remain in the literature regarding sequential adaptation using modern PEFT techniques for LLMs. In particular, the impact of task ordering on knowledge transfer, catastrophic forgetting, and performance trade-offs has not been systematically investigated. This study addresses this gap through an empirical analysis of sequential PEFT-based adaptation for mental health NLP tasks. Beyond the current perspective, this study also examines sequential PEFT as a computationally efficient approach to adapt large language models under constrained computational resources, thus connecting task-order optimization with the broader high-performance computing HPC objective of maximizing model performance relative to computational and memory costs.

# 3. Methodology

This study investigates the efficacy of sequential MTL using PEFT to adapt the Llama 3.1 8B Instruct model for mental health problem detection from text, leveraging two auxiliary tasks. Figure 2 summarizes the complete experimental workflow, beginning with dataset preparation and preprocessing, followed by PEFT-based adaptation of Llama 3.1 8B Instruct and sequential multi-task fine-tuning under two task-ordering curricula. The tasks are defined as follows.

- Mental Health Problem (MHP) Detection (Primary Task): Classifying textual reviews (social media posts) into Anxiety, Bipolar, Depression, or SuicideWatch classes.
- Multi-Label Emotion Detection (Auxiliary Task): Identifying the presence of eight emotions (anger, fear, hopefulness, hopelessness, joy, sadness, calmness, disgust) in reviews.

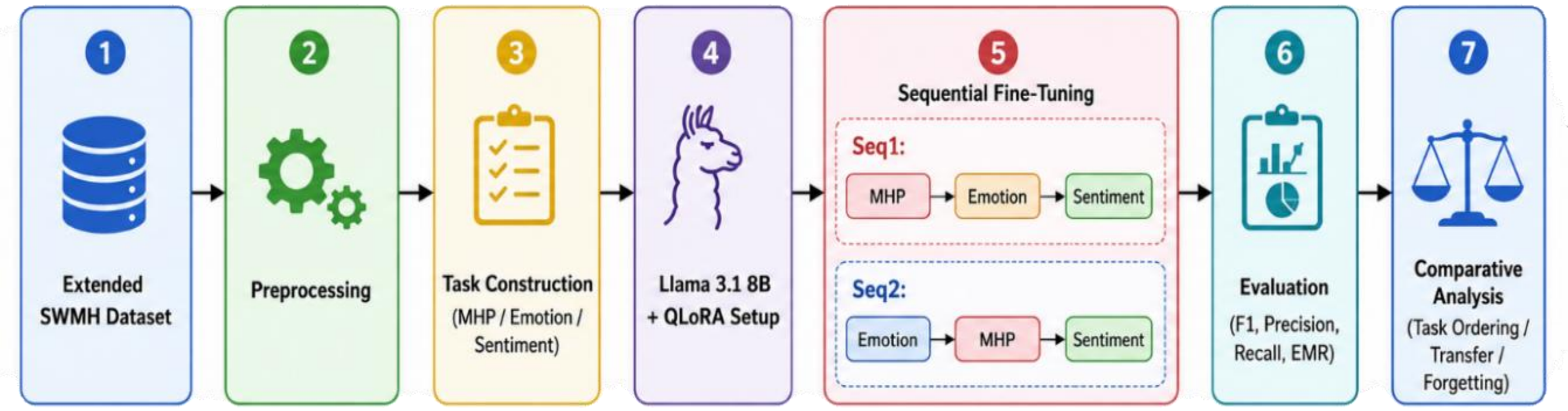

Figure 2: Overall methodology and sequential multi-task learning workflow

- Sentiment Analysis (Auxiliary Task): Classifying the reviews as Positive or Negative. Due to the extremely limited number of neutral instances in the dataset, the sentiment task was formulated as a binary classification problem to ensure stable model training. Specifically, during the programmatic data preparation for the binary sentiment task, these 588 neutral instances (where explicit Positive and Negative annotations were 0) were deterministically mapped to the 'Negative' class. This mapping decision reflects the fact that unstructured posts sourced from clinical mental health subreddits overwhelmingly exhibit underlying negative affective polarity, allowing us to maintain the full 4,243-instance training volume without artificially introducing a tertiary 'neutral' class.

Our methodology encompasses dataset preparation, model and PEFT configuration, Sequential MTL experiment design, and evaluation phases. The following subsections explain each phase in detail.

### *3.1.* Research Questions

This study aims to answer the following research questions:

- RQ1 (Primary Task Enhancement): Does sequentially fine-tuning a PEFT-adapted LLM on related auxiliary tasks improve its performance on a primary Mental Health Problem Detection task?
- RQ2 (Auxiliary Task Dynamics & Transfer): How does sequential fine-tuning on a new auxiliary task impact the performance of previously learned auxiliary tasks?
- RQ3 (Comparative Performance & Efficiency): How does the performance of a sequentially fine-tuned Llama 3.1 8B model using PEFT compare to its base zero-shot performance?
- RQ4 (Limitations and Challenges): What are the observable limitations, such as task interference, encountered when employing this sequential MTL approach?

### *3.2 Dataset*

Our research utilizes the extended SWMH (SuicideWatch and Mental Health) dataset, which contains Reddit posts relevant to various Mental health problems. Armah & Dehkharghani (2024) [4] manually augmented 5,000 instances of the dataset with sentiment and emotion annotations and made it publicly available on GitHub [26].

Table I presents four sample reviews from the extended SWMH dataset, showing the multi-labeled structure in which each text instance is annotated for mental health problems, emotions, and sentiment polarity.

Table I: An example from the extended SWMH dataset.

| Statement | Mental Health Problem | Sentiment | | Emotions | | | | | | | |
|---|---|---|---|---|---|---|---|---|---|---|---|
| | | Pos | Neg | anger | fear | hopefulness | hopelessness | joy | sadness | calmness | disgust |
| wanting to skip or postpone my exam my exam is on saturday and i feel a massive urge to reschedule i experience this often with major exams. *flashbacks to several years prior where i skipped all of my final exams and failed* i just dont feel ready i dont know im so tired and mentallly fried that i feel a few days might help. but a part of me thinks i should fight this urge and just tank it | Anxiety | 0 | 1 | 0 | 1 | 0 | 1 | 0 | 0 | 0 | 0 |
| Anonymous Entry: plz be nice My brain boils non-stop lol. I've finally become the deteriorated husk of a person. Hopefully this is my last feeling of self-awareness. | depression | 0 | 1 | 0 | 0 | 0 | 1 | 0 | 1 | 0 | 0 |
| I made a quilt! I formulated a plan. On another day, I ordered materials. A different day, I made the front of the quilt. A couple of weeks later I shopped for a good back of quilt fabric. A week later, I did all the fancy sewing to attach the back to the front, and today I did the binding!!! I did a project and didn't try to do it all at once. And I FINISHED it!!! | bipolar | 1 | 0 | 0 | 0 | 0 | 0 | 1 | 0 | 1 | 0 |
| I'm gonna do it tonight I've attempted 4 times already. Why not once more? | SuicideWatch | 0 | 1 | 0 | 0 | 0 | 1 | 0 | 0 | 0 | 0 |

Figures 3-5 show the class distributions for the three tasks: mental health, emotion, and sentiment detection.

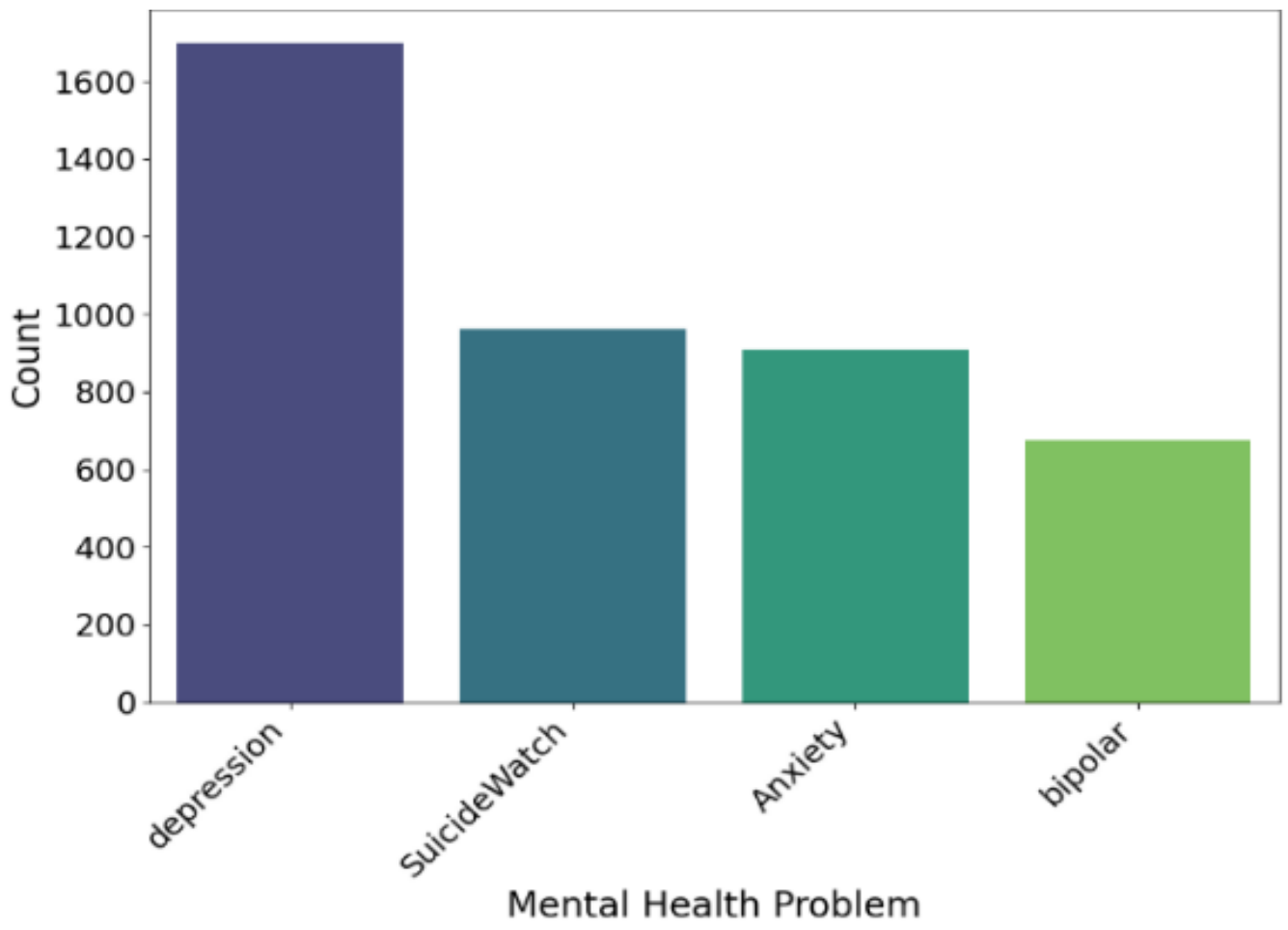

Figure 3: Distribution of classes for Mental Health Problem detection

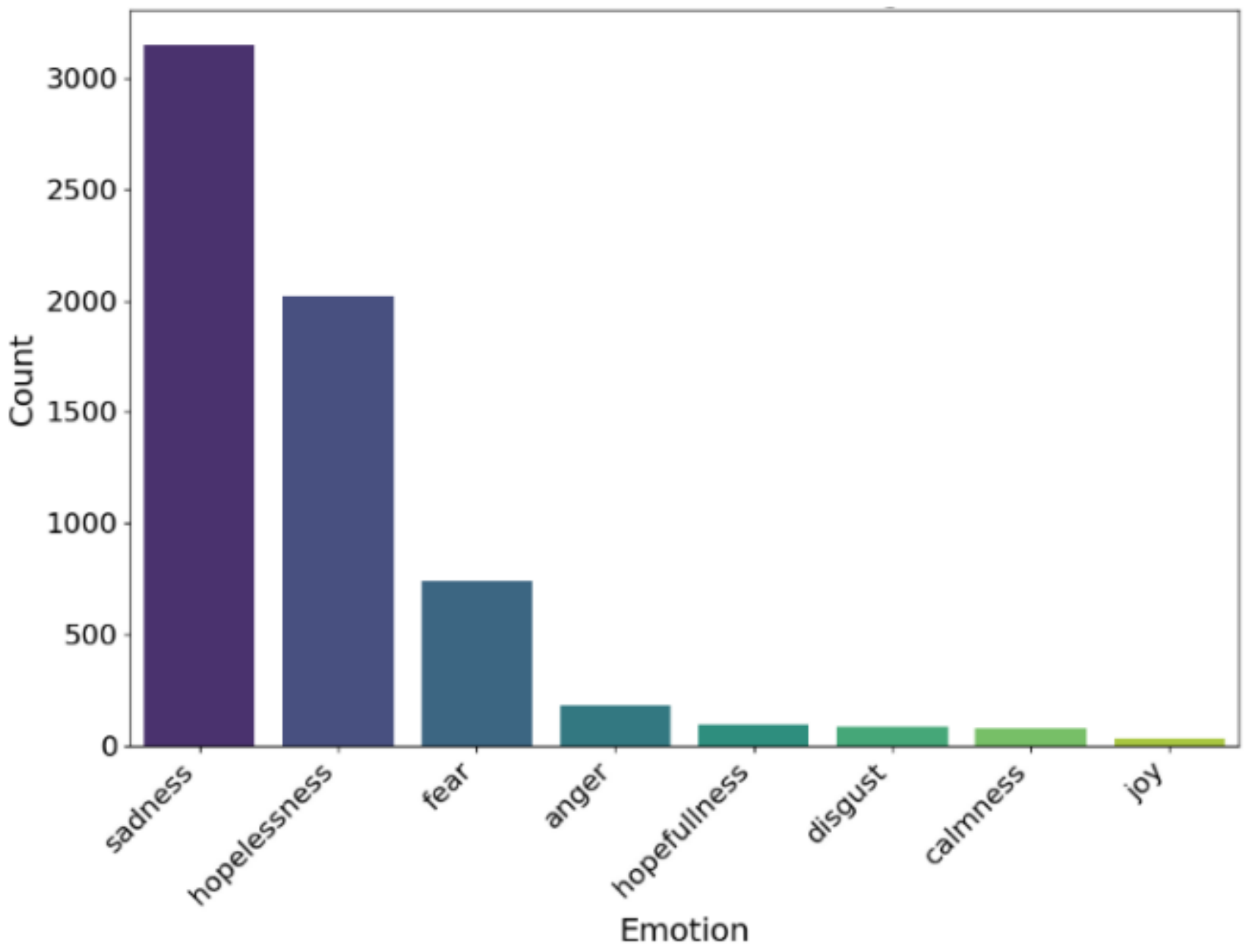

Figure 4: Distribution of classes for Emotion detection

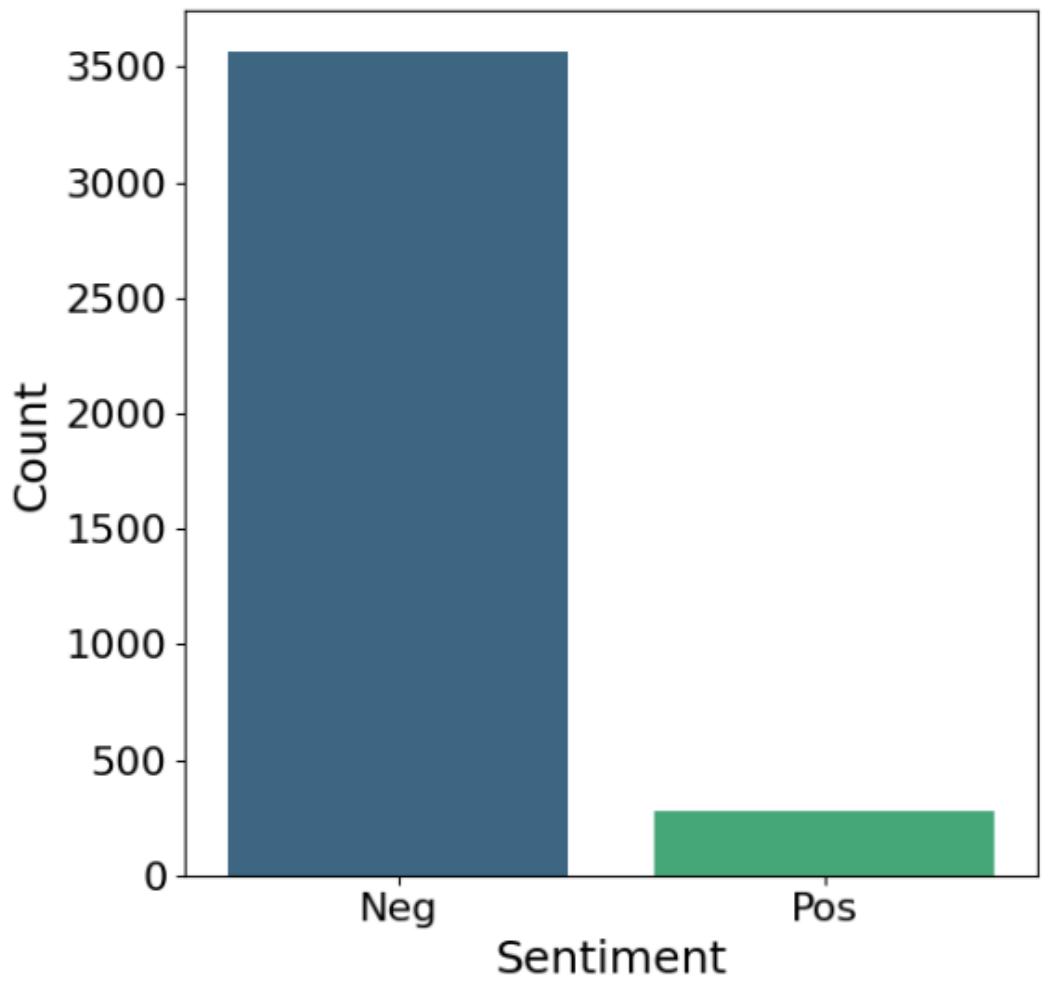

Figure 5: Distribution of classes Sentiment detection

These figures illustrate the significant class imbalance, particularly for minority emotion classes and the positive sentiment class. To further justify the use of MTL, Figure 6 presents the task correlation matrix to analyze potential task-relatedness. As an outstanding example, classes such as ‘depression positively correlate with 'hopelessness' and 'sadness' but negatively with ‘fear’.

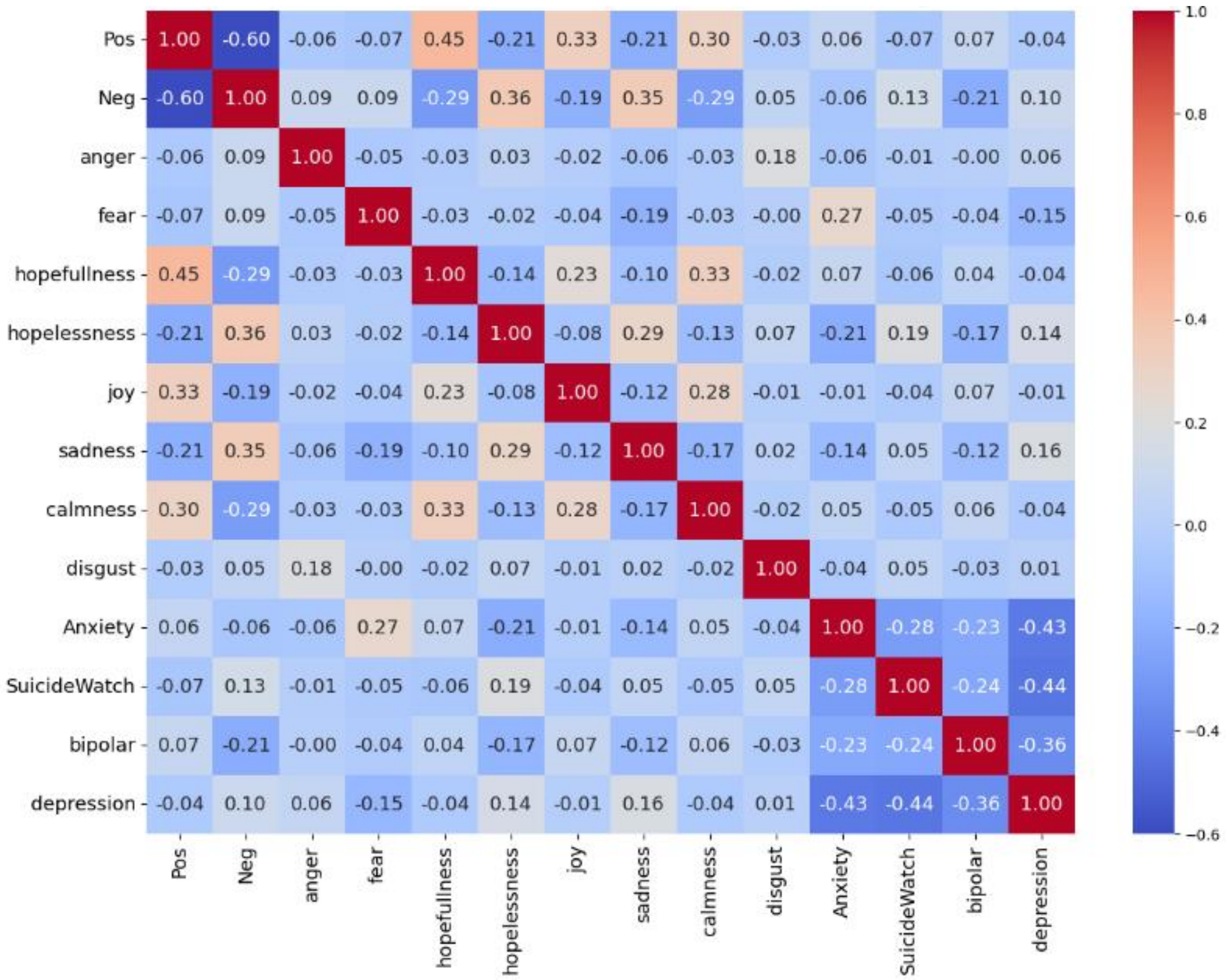

Figure 6: Correlation heatmap of all labels in the dataset.

### *3.3. Data Preprocessing*

A critical preprocessing step involved refining the MHP task labels. The original SWMH data included an 'offmychest' category (from the r/offmychest subreddit). Preliminary experiments indicated that including 'offmychest' as a target "disorder" confused the LLM due to its semantic incongruity and content heterogeneity, hindering the model's learning. Consequently, all 'offmychest' instances were excluded from the dataset used for MHP detection and all subsequent fine-tuning stages. This resulted in a curated dataset of 4,243 records for this study. The emotion and sentiment annotations for these remaining records were retained. To preserve the natural distribution of the social media posts, the dataset was divided using a sequential 80:10:10 split (rather than stratified sampling) into training (3,394 instances), validation (425 instances), and test (425 instances) sets.

All text data underwent a standard NLP preprocessing pipeline: lowercasing, removal of URLs and HTML tags, removal of non-alphabetic characters, tokenization (NLTK word_tokenize), stopword removal (NLTK list), and WordNet lemmatization (NLTK WordNetLemmatizer). The cleaned statement was used as input to the LLM.

To assess the impact of this preprocessing pipeline on model performance, we also conducted parallel experiments using raw, unprocessed text. While the impact on predictive performance was minimal, the preprocessing pipeline was deliberately retained to remove noisy social media artifacts (URLs, HTML), ensure clean tokenization, and maintain strict methodological alignment with established NLP benchmarks.

***3.4. Model and Parameter-Efficient Fine-Tuning (PEFT)***

- Base Model: We employed the meta-llama/Llama-3.1-8B-Instruct model [2], a decoder-only Transformer with approximately 8 billion parameters, 32 layers, a model dimension of 4096, 32 attention heads (8 key/value heads for GQA), an FFN dimension of 14336, and a 128K vocabulary. It utilizes RoPE positional embeddings and the SwiGLU activation function. This instruction-tuned model was accessed via the Hugging Face transformers library [33].
- PEFT with LoRA and QLoRA: To adapt the LLM efficiently, we used Low-Rank Adaptation (LoRA) [3] with 4-bit quantization based on QLoRA principles [16]. Table II provides the models configurations used.

Table II: Model Configurations.

| **Category** | **Parameter** | **Value / Setting** |
|---|---|---|
| Quantization Settings | load_in_4bit | True |
| | bnb_4bit_quant_type | "nf4" |
| | bnb_4bit_use_double_quant | True |
| | bnb_4bit_compute_dtype | torch.float16 |
| LoRA Adapter Settings | r (Rank) | 16 |
| | lora_alpha | 16 |
| | lora_dropout | 0.1 |
| | bias | "none" |
| | task_type | "CAUSAL_LM" |
| | Target Modules | All linear layers (bnb.nn.Linear4bit), excluding the language model head. |

This approach trains less than 1% of the total model parameters, enabling efficient fine-tuning. This significantly reduces GPU memory consumption and computational overhead compared to full-model fine-tuning, enabling practical adaptation of LLMs with limited hardware resources while preserving competitive downstream performance. Despite the reduction in trainable parameters, the underlying 8-billion-parameter Transformer remains computationally intensive because the frozen base model must still participate in forward and backward computations during training.

Hyperparameter settings: The hyperparameter values outlined in Table III were strategically selected based on a combination of standard LLM adaptation practices, dataset characteristics, and strict computational resource constraints.

Table III: Task-specific fine-tuning hyperparameters

| **Hyperparameter** | **MHP Task** | **Emotion Task** | **Sentiment Task** |
|---|---|---|---|
| num_train_epochs | 10 | 3 | 3 |
| per_device_train_batch_size | 1 | 2 | 2 |
| gradient_accumulation_steps | 16 | 8 | 8 |
| learning_rate | 5e-5 | 3e-5 | 3e-5 |
| weight_decay | 0.01 | 0.03 | 0.03 |

The justification of the chosen values for parameters is discussed below:

1. LoRA Parameters (Rank and Alpha): Both r and lora_alpha were set to 16. This provides a proven balance between maintaining parameter efficiency (training on less than 1% of the weights) and retaining sufficient representational capacity to learn the complex linguistic nuances of mental health text.
2. Batch Size and Gradient Accumulation: Due to the severe GPU memory limitations inherent in fine-tuning an 8-billion parameter model (even with 4-bit quantization on NVIDIA Tesla T4/P100 GPUs), the per-device batch size was restricted to 1 or 2. To compensate and ensure stable gradient updates, gradient accumulation steps were set to 16 or 8, respectively, achieving an effective batch size of 16 across all tasks.
3. Epochs and Early Stopping: The primary task (MHP Detection) was fine-tuned for a maximum of 10 epochs to establish a strong foundational representation of the clinical classes. Conversely, the auxiliary tasks (Emotion and Sentiment) were restricted to 3 epochs. This deliberate reduction, paired with an Early Stopping callback (patience=3), was implemented to prevent overfitting on the highly imbalanced emotion dataset and to mitigate task interference affecting the primary MHP representations during the sequential multi-task learning process.

### *3.5. Sequential Multi-Task Learning Experimental Design*

We implemented and evaluated two different sequential fine-tuning curricula. Seq1: MHP-First (Primary), and Seq2: Emo-First (Alternate), which are illustrated in Figure 7 and Figure 8. Evaluating all six possible task permutations would provide a more exhaustive characterization of task-order effects. However, our experimental design intentionally focused on two contrasting and theoretically informative curricula rather than attempting to identify a globally optimal ordering through exhaustive permutation search.

Specifically, Seq1 (MHP → Emotion → Sentiment) was selected as a primary-task-first curriculum, in which the target task, Mental Health Problem (MHP) detection, is learned before the two related auxiliary tasks. Seq2 (Emotion → MHP → Sentiment) was selected as an auxiliary-task-first curriculum, in which the model first learns affective representations from the Emotion task and then adapts to the primary MHP task. Thus, the two sequences directly contrast two distinct curriculum strategies: prioritizing the primary task from the beginning versus first acquiring knowledge from a related auxiliary task. This design allows us to directly examine whether placing the primary MHP task at the beginning or after an auxiliary task leads to different performance trade-offs across the three tasks.

We acknowledge that evaluating the remaining four permutations would provide a more comprehensive characterization of the task-order space. However, sequentially fine-tuning an 8-billion-parameter LLM makes exhaustive permutation evaluation computationally demanding within our resource-constrained HPC environment. Therefore, rather than arbitrarily selecting two permutations, we deliberately selected these two contrasting curricula because they directly address the central comparison investigated in this study.

Furthermore, to strictly isolate the impact of task-ordering on knowledge transfer, we utilized standard cross-entropy loss without implementing class-balancing interventions (e.g., focal loss or resampling), as such techniques could confound the sequential transfer dynamics.

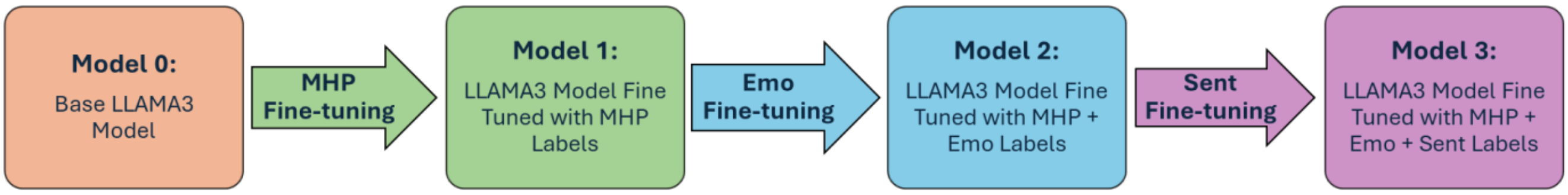

Figure 7: Sequential stages of fine-tuning process of Seq1: MHP→Emo→Sent.

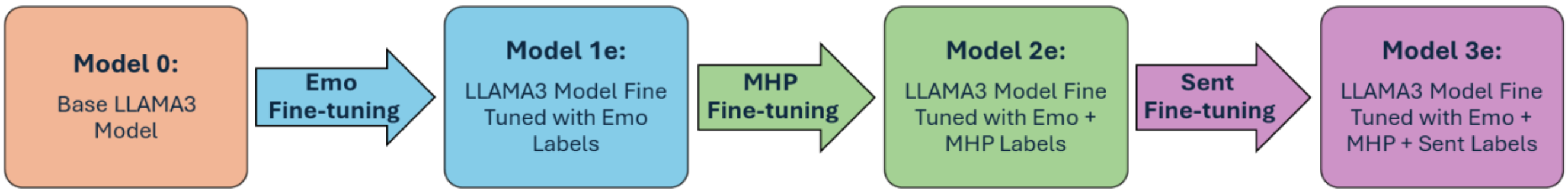

Figure 8: Sequential stages of fine-tuning process of Seq2: Emo→MHP→Sent.

1. Sequence 1 (MHP-first: M0 → M1 → M2 → M3):
    - M0 (Base): Unmodified Llama 3.1 8B Instruct (quantized).
    - M1 (MHP-Tuned): M0 fine-tuned on MHP Detection.
    - M2 (MHP+Emo-Tuned): M1 further fine-tuned on Emotion Detection.
    - M3 (MHP+Emo+Sent-Tuned): M2 further fine-tuned on Sentiment Analysis.

2. Sequence 2 (Emo-first: M0 → M1e → M2e → M3e):

- M0 (Base): Unmodified Llama 3.1 8B Instruct (quantized).
- M1e (Emo-Tuned): M0 fine-tuned on Emotion Detection.
- M2e (Emo+MHP-Tuned): M1e further fine-tuned on MHP Detection.
- M3e (Emo+MHP+Sent-Tuned): M2e further fine-tuned on Sentiment Analysis.

For subsequent stages in each sequence, the PEFT model from the previous stage (frozen base LLM + trained LoRA adapters) was loaded, and the SFTTrainer from the trl library continued to update these existing adapter weights with the new task's data.

#### *3.5.1. Prompt Engineering*

For generative, instruction-tuned LLMs, natural language prompting serves as the required native interface rather than an optional enhancement. Instruction-based prompts were chosen over alternative approaches such as few-shot prompting or prompt tuning because they align naturally with the instruction-following behavior of the Llama 3.1 8B Instruct model. To empirically verify the baseline effectiveness of these prompt templates, the evaluation of the Zero-Shot base model (M0) serves as an inherent ablation study, demonstrating the model's predictive capability utilizing the prompts alone prior to any parameter adaptation. By framing each task as an explicit natural language instruction, the prompts leverage the model's pre-trained instruction-following capabilities, thereby reducing the risk of task misinterpretation during both fine-tuning and inference. This approach also ensures a consistent interface across all three tasks, making the sequential fine-tuning pipeline easier to manage and reproduce. Output formatting constraints were enforced directly within the prompt text: for MHP detection, the model was instructed to return exactly one class label (Anxiety, Bipolar, Depression, or SuicideWatch); for multi-label emotion detection, it was asked to return a comma-separated list of applicable emotion labels; and for sentiment analysis, it was constrained to a single binary label (Positive or Negative). These explicit output specifications reduce post-processing complexity and minimize the risk of the model generating verbose or free-form responses that require additional parsing.

To ensure consistency across tasks, all three prompts share the same structural template: a concise task instruction, the input field referencing the preprocessed statement, and a label field that is populated during fine-tuning but left blank during inference. This uniform structure facilitates a fair comparison of task-order effects, since the model is exposed to the same prompt format at every sequential stage. Prompt simplicity was a deliberate design choice: overly complex or lengthy prompts risk introducing spurious contextual cues that could confound the model's predictions, particularly for the sensitive mental health domain. Simple, unambiguous instructions reduce cognitive load on the model, help maintain stable gradient behavior during fine-tuning, and make the experimental setup more transparent and reproducible for future research. During the inference stage, generation was performed with a temperature of 0.1 to enforce deterministic, strictly formatted outputs that adhered to the predefined labels.

The details of the prompts are provided for reproducibility purposes.

1. Mental Health Problem Detection Prompt:

   *Classify the text into Anxiety, bipolar, depression, SuicideWatch, and return the answer as the corresponding mental health problem label.*

   *text: {data_point["cleaned_statement"]}*

   *label: {data_point["status"]}*

2. Multi-Label Emotion Detection Prompt:

   *Classify the following text into one or more emotion categories: anger, fear, hopefulness, hopelessness, joy, sadness, calmness, disgust. Return a comma-separated list of emotions that apply to the text.*

   *Text: {data_point["cleaned_statement"]}*

   *Emotions: {data_point["emotion_labels"]}*

3. Sentiment Analysis Prompt:

   *Classify the following text as either Positive or Negative sentiment.*

   *text: {data_point["cleaned_statement"]}*

   *sentiment: {data_point["sentiment"]}*

4. Joint Multi-Task Learning (Joint-MTL) Unified Prompt:

   *Analyze the following text and perform three tasks: 1. Classify the mental health disorder (Anxiety, bipolar, depression, SuicideWatch). 2. Extract all applicable emotions (anger, fear, hopefulness, hopelessness, joy, sadness, calmness, disgust). 3. Classify the sentiment (Positive or Negative).*

   *text: {data_point["cleaned_statement"]}*

   *Output: MHP: {data_point["status"]}*

   *Emotion: {emotions_str}*

   *Sentiment: {sentiment_str}*

#### *3.5.2. Fine-tuning Configuration*

All fine-tuning stages were managed using the SFTTrainer from the trl library with TrainingArguments. Consistent settings included. Table IV provides the parameters with their values for this configuration. Note that evaluation was performed every 50 or 100 steps, and experiments were tracked using WandB.

Table IV: Fine-tuning parameter configuration.

| Parameter | Value |
|---|---|
| optim | "paged_adamw_32bit" |
| fp16 | True |
| gradient_checkpointing | True |
| lr_scheduler_type | "cosine" |
| load_best_model_at_end | True (based on eval_loss) |
| EarlyStoppingCallback(patience) | 3 |

### *3.6. Evaluation Metrics and Baseline*

Model performance was evaluated on the test set for all three tasks after each fine-tuning stage using standard classification evaluation metrics. For MHP and Sentiment detection (single-label), we report Precision, Recall, and F1-score (per-class, macro, micro, weighted). For Emotion detection (multi-label), we report per-emotion binary metrics and overall Exact Match Ratio (EMR), Micro-F1, and Macro-F1. The Micro-averaged F1-score is used as the primary comparative metric. All evaluation metrics are calculated and reported with exact, unrounded four-decimal precision to ensure rigorous comparison. To quantify test-set uncertainty, we calculated 95% confidence intervals(CI) for the primary Micro-F1 scores using the standard normal approximation (Wald interval). For the single-label tasks (MHP and Sentiment), N=425 test instances were used; for the multi-label Emotion task, N=599 total true positive support labels were used. Moreover, to establish robust performance benchmarks, model evaluation is structured around three comparative axes: (1) the zero-shot capabilities of the unadapted Llama 3.1 8B Instruct model (M0); (2) isolated single-task fine-tuning, represented by the first stage of each sequential pipeline (M1 for MHP and M1e for Emotion); and (3) the fully realized sequential multi-task models (M3 and M3e). To rigorously isolate the efficacy of our proposed sequential task-ordering methodology, we also established a Joint Multi-Task Learning (Joint-MTL) Baseline. The Llama-3.1-8B-Instruct model was fine-tuned to predict MHP, Emotion, and Sentiment simultaneously using a single, unified prompt. Instead of isolating tasks sequentially, this "prompt packing" strategy required the model to generate all three classifications at once from a single input sequence. The same QLoRA hyperparameters, base model, and training epochs were used to ensure a fair, controlled comparison. The Mj0 model evaluated on this unified prompt serves as our Zero-Shot Joint Baseline.

# 4. Experimental Results

This section presents the empirical results from the sequential multi-task fine-tuning of the Llama 3.1 8B Instruct model. We first provide an overview of model performance for both primary and alternate fine-tuning sequences (MHP→Emo→Sent and Emo→MHP→Sent). Then, we analyze the impact of auxiliary-task fine-tuning on the primary Mental Health Problem Detection task (RQ1), examine the performance dynamics of auxiliary tasks (RQ2), compare against baselines (RQ3), and discuss observed limitations (RQ4).

### *4.1. Overview of Model Performance Across Sequential Stages*

Across sequential fine-tuning stages, the Llama 3.1 8B Instruct model's detection performance on each task (MHP, Emotion, and Sentiment) was evaluated using the Micro-F1 score. This metric is particularly effective for assessing a model on an imbalanced dataset, as in our case. The core performance of the Llama 3.1 8B Instruct model, from its baseline zero-shot capabilities through the final stage of both sequential fine-tuning curricula, is summarized in Table V. This table presents the Micro-F1 scores for all individual tasks. Seq1 and Seq2 refer to different task orders. M0 is the base model; M1/M1e are models after the first stage; M2/M2e are models after the second; and M3/M3e are the final models after being finetuned across all three tasks.

Table V: Micro F1-Scores Across Sequential Fine-Tuning Stages

| **Model / Sequence** | **MHP Detection** | **Emotion Detection** | **Sentiment Detection** |
|---|---|---|---|
| **M0: LLAMA3 Base Model (Zero-Shot)** | 0.6188 | 0.2930 | 0.5647 |
| **Seq1: MHP→Emo→Sent** | | | |
| **M1: MHP-Tuned** | 0.7388 **(+0.1200)** | 0.4663 **(+0.1733)** | 0.8635 **(+0.2988)** |
| **M2: MHP+Emo-Tuned** | 0.7553 **(+0.0165)** | 0.4697 **(+0.0034)** | 0.8635 **(+0.0000)** |
| **M3: MHP+Emo+Sent-Tuned (Final Seq1)** | 0.7624 **(+0.0071)** | 0.4385 **(-0.0312)** | 0.8635 **(+0.0000)** |
| **Seq2: Emo→MHP→Sent** | | | |
| **M1e: Emo-Tuned** | 0.6259 **(+0.0071)** | 0.6004 **(+0.3074)** | 0.9459 **(+0.3812)** |
| **M2e: Emo+MHP-Tuned** | 0.6824 **(+0.0565)** | 0.5978 **(-0.0026)** | 0.9388 **(-0.0071)** |
| **M3e: Emo+MHP+Sent-Tuned (Final Seq2)** | 0.6259 **(-0.0565)** | 0.6052 **(+0.0074)** | 0.9459 **(+0.0071)** |

The results show that both sequential fine-tuning strategies improved the performance over the base Llama 3.1 (M0). However, the final outcomes for each task varied notably depending on the fine-tuning sequence. Tables VI and VII present the detailed classification report, including Precision, Recall, and F1-scores for the test set of the fine-tuned Llama model, separately for Seq1 and Seq2, across all tasks.

Table VI: Detailed Classification Report for the Final Model (M3) of the MHP→Emo→Sent Sequence (Seq1) on all tasks

| | Precision | Recall | F1-Score | Support |
|---|---|---|---|---|
| **Task 1: Mental Health Problem Detection** | | | | |
| Anxiety | 0.8409 | 0.8222 | 0.8315 | 90 |
| SuicideWatch | 0.6931 | 0.6630 | 0.6778 | 92 |
| bipolar | 0.9230 | 0.7500 | 0.8276 | 80 |
| depression | 0.7010 | 0.7914 | 0.7435 | 163 |
| Micro avg | 0.7624 | 0.7624 | 0.7624 | 425 |
| Macro avg | 0.7895 | 0.7566 | 0.7701 | 425 |
| Weighted avg | 0.7707 | 0.7624 | 0.7635 | 425 |
| **Task 2: Emotion Detection** | | | | |
| anger | 0.1892 | 0.6364 | 0.2917 | 22 |
| fear | 0.3506 | 0.3971 | 0.3724 | 68 |
| hopefulness | 0.0000 | 0.0000 | 0.0000 | 4 |
| hopelessness | 0.4316 | 0.2193 | 0.2908 | 187 |
| joy | 0.0000 | 0.0000 | 0.0000 | 2 |
| sadness | 0.7750 | 0.4968 | 0.6055 | 312 |
| calmness | 0.0000 | 0.0000 | 0.0000 | 2 |
| disgust | 0.0000 | 0.0000 | 0.0000 | 2 |
| Micro avg | 0.4917 | 0.3957 | 0.4385 | 599 |
| Macro avg | 0.2183 | 0.2187 | 0.1950 | 599 |
| **Task 3: Sentiment Detection** | | | | |
| Negative | 0.9802 | 0.8696 | 0.9216 | 399 |
| Positive | 0.2816 | 0.7692 | 0.4123 | 26 |
| Micro avg | 0.8635 | 0.8635 | 0.8635 | 425 |
| Macro avg | 0.6309 | 0.8194 | 0.6669 | 425 |
| Weighted avg | 0.9374 | 0.8635 | 0.8904 | 425 |

### *4.2. Impact of Auxiliary Task Fine-tuning on Primary Task (RQ1)*

Our primary goal was to optimize the detection of Mental Health Problems (MHPs). The primary MHP-first sequence (Seq1) demonstrated a clear and continuous benefit from sequential fine-tuning. As shown in Table V, initial fine-tuning on the MHP task (M1) yielded a substantial performance increase from 0.6188 to 0.7388 Micro F1. Subsequent fine-tuning on emotion (M2) and sentiment (M3) tasks provided further incremental gains, culminating in a final Micro F1 score of 0.7624. This confirms that learning-related auxiliary tasks provide a positive transfer for the primary clinical task.

Table VII: Detailed Classification Report for the Final Model (M3e) of the Emo→ MHP→Sent Sequence (Seq2) on all tasks

| | **Precision** | **Recall** | **F1-Score** | **Support** |
|---|---|---|---|---|
| **Task 1: Emotion Detection** | | | | |
| anger | 0.1287 | 0.5909 | 0.2114 | 22 |
| fear | 0.3934 | 0.3529 | 0.3721 | 68 |
| hopefulness | 0.1053 | 0.5000 | 0.1739 | 4 |
| hopelessness | 0.5072 | 0.9358 | 0.6579 | 187 |
| joy | 0.1429 | 0.5000 | 0.2222 | 2 |
| sadness | 0.7815 | 0.8942 | 0.8341 | 312 |
| calmness | 0.0105 | 0.5000 | 0.0206 | 2 |
| disgust | 0.0182 | 0.5000 | 0.0351 | 2 |
| Micro avg | 0.4769 | 0.8280 | 0.6052 | 599 |
| Macro avg | 0.2610 | 0.5967 | 0.3159 | 599 |
| **Task 2: Mental Health Problem Detection** | | | | |
| Anxiety | 0.8166 | 0.5444 | 0.6533 | 90 |
| SuicideWatch | 0.6911 | 0.5108 | 0.5875 | 92 |
| bipolar | 0.8823 | 0.3750 | 0.5263 | 80 |
| depression | 0.5323 | 0.8588 | 0.6572 | 163 |
| micro avg | 0.6259 | 0.6259 | 0.6259 | 425 |
| macro avg | 0.7305 | 0.5722 | 0.6060 | 425 |
| weighted avg | 0.6927 | 0.6259 | 0.6163 | 425 |
| **Task 3: Sentiment Detection** | | | | |
| Negative | 0.9676 | 0.9749 | 0.9712 | 399 |
| Positive | 0.5652 | 0.5000 | 0.5306 | 26 |
| micro avg | 0.9459 | 0.9459 | 0.9459 | 425 |
| macro avg | 0.7664 | 0.7374 | 0.7509 | 425 |
| weighted avg | 0.9430 | 0.9459 | 0.9442 | 425 |

Conversely, when MHP detection was the second task in the alternate Emo-first sequence (Seq2), the final performance was significantly lower at 0.6259 Micro F1. Intriguingly, performance in this sequence peaked after MHP tuning (M2e) and then slightly degraded after the final sentiment tuning stage, suggesting negative interference. This contrast suggests that placing the primary task first may provide an advantage for MHP detection within the evaluated task-order configurations.

### *4.3. Impact of Sequential Tuning on Auxiliary Task Performance (RQ2)*

The performance trajectories of the auxiliary tasks revealed a more complex interplay of knowledge transfer and interference.

- Emotion Detection: The Emo-first sequence (Seq2) was clearly superior for this task, achieving a final Micro F1 of 0.6052. In contrast, the MHP-first sequence (Seq1) showed a strong initial boost from MHP tuning but ultimately degraded after sentiment tuning, finishing with a much lower score of 0.4385. This suggests that while MHP features are helpful for emotion detection, subsequent optimization for a simpler task, such as sentiment, can interfere with and undermine the complex, multi-label representations needed for emotion detection. Notably, across both sequences, the model struggled with highly imbalanced classes like 'hopefulness' and 'joy', often failing to predict them entirely

- Sentiment Detection: The Emo-first sequence (Seq2) also proved optimal for sentiment analysis, achieving an exceptionally high Micro F1 of 0.9459. The initial emotion tuning provided a massive performance jump, which was largely sustained throughout the subsequent stages. In the MHP-first sequence (Seq1), sentiment performance also improved dramatically after the first stage but plateaued early at 0.8635, with no further gains from direct sentiment tuning. This indicates that the features learned from either emotion or Mental health problems were already sufficient to nearly master the simpler binary sentiment task.

***4.4. Comparative Performance Against Baselines (RQ3)***

Table IX provides a comprehensive overview of model performance across all tasks, explicitly contrasting two distinct learning paradigms: simultaneous Joint-MTL and our proposed Sequential Fine-Tuning. To fully evaluate the efficacy of our sequential methodology, we contextualize these results using three comparative axes:

- **Zero-Shot LLM (M0):** The final sequentially fine-tuned models substantially outperformed the unmodified Llama 3.1 base model across all tasks. As shown in Table IX, the MHP-first sequence (M3) achieved a Micro-F1 of 0.7624 for MHP detection, representing a substantial gain of +0.1436 over the zero-shot baseline (0.6188).

- **Single-Task PEFT Baselines (M1 & M1e):** The initial stage of our training pipelines effectively isolates single-task adaptation. Model M1, representing the LLM fine-tuned exclusively on the MHP task, achieved a **0.7388** Micro-F1. By the conclusion of the sequential pipeline (M3), this performance increased to **0.7624.** Similarly, model M1e isolates single-task training for Emotion, achieving **0.6004** Micro-F1; by the end of its sequence (M3e), performance further improved to **0.6052**. This progression across both tasks demonstrates that the sequential introduction of related auxiliary tasks provides a measurable enhancement over isolated single-task adaptation.

- **Joint-MTL and Prior Architectures:** While prior research utilizing this dataset (Armah & Dehkharghani, 2024) explored Joint MTL via custom neural architectures (Conv1D/BiLSTM), a direct one-to-one numerical comparison is not possible. As detailed in Section 3.3, we intentionally excluded the heterogeneous 'offmychest' class to ensure semantic coherence for LLM instruction tuning. This shifted the primary evaluation from a 5-class to a curated 4-class problem. However, to provide a strictly controlled, apples-to-apples comparison, we directly compared our sequential framework against our own in-domain simultaneous Joint-MTL baselines (**Table IX**). As highlighted by the structural grouping in the table, the Joint-MTL paradigm significantly underperforms our sequential approach. When evaluated in a zero-shot joint setting (Mj0), the base model suffered complete generative format collapse (e.g., 0.0094 MHP F1), performing substantially worse than individual zero-shot prompting (M0). When fine-tuned, the Trained Joint-MTL model (Mj) improved formatting but suffered from severe task interference. While it allocated sufficient capacity to predict Sentiment (0.7741), its performance on the complex MHP task catastrophically collapsed to a Micro-F1 of 0.1953. This is a textbook manifestation of negative transfer. Furthermore, the cognitive overload caused the Mj model to frequently hallucinate invalid categorical labels or omit labels entirely (captured as "None" in our evaluation reports; see **Table VIII** below). In stark contrast, our Sequential Fine-Tuning methodology achieved 100% format compliance with zero invalid categorical outputs, empirically validating that sequential task isolation can be advantageous for the evaluated generative LLM configuration.

Table VIII: Detailed Classification Report for the Trained Joint-MTL Baseline (Mj) across all tasks, highlighting instances of generative collapse (hallucinated 'None' labels).

| | **Precision** | **Recall** | **F1-Score** | **Support** |
|---|---|---|---|---|
| **Task 1: Mental Health Problem Detection** | | | | |
| Anxiety | 0.2128 | 0.9222 | 0.3458 | 90 |
| SuicideWatch | 0.0000 | 0.0000 | 0.0000 | 92 |
| bipolar | 0.0000 | 0.0000 | 0.0000 | 80 |
| depression | 0.0000 | 0.0000 | 0.0000 | 163 |
| None (Hallucinated) | 0.0000 | 0.0000 | 0.0000 | 0 |
| Micro avg | 0.1953 | 0.1953 | 0.1953 | 425 |
| Macro avg | 0.0426 | 0.1844 | 0.0692 | 425 |
| Weighted avg | 0.0451 | 0.1953 | 0.0732 | 425 |
| **Task 2: Emotion Detection** | | | | |
| anger | 0.1833 | 0.5000 | 0.2683 | 22 |
| fear | 0.5217 | 0.3529 | 0.4211 | 68 |
| hopefulness | 0.2000 | 0.2500 | 0.2222 | 4 |
| hopelessness | 0.4579 | 0.6684 | 0.5435 | 187 |
| joy | 0.0000 | 0.0000 | 0.0000 | 2 |
| sadness | 0.8341 | 0.6122 | 0.7061 | 312 |
| calmness | 0.0000 | 0.0000 | 0.0000 | 2 |
| disgust | 0.0351 | 1.0000 | 0.0678 | 2 |
| Micro avg | 0.5050 | 0.5910 | 0.5446 | 599 |
| Macro avg | 0.2790 | 0.4229 | 0.2786 | 599 |
| **Task 3: Sentiment Detection** | | | | |
| Negative | 0.9347 | 0.8246 | 0.8762 | 399 |
| Positive | 0.0000 | 0.0000 | 0.0000 | 26 |
| None (Hallucinated) | 0.0000 | 0.0000 | 0.0000 | 0 |
| Micro avg | 0.7741 | 0.7741 | 0.7741 | 425 |
| Macro avg | 0.3116 | 0.2749 | 0.2921 | 425 |
| Weighted avg | 0.8775 | 0.7741 | 0.8226 | 425 |

- **Detailed Joint-MTL Performance and Generative Collapse:** Table VIII presents the detailed class-level classification reports for the Trained Joint-MTL baseline (Mj). As shown, the cognitive overload of simultaneous multitasking resulted in the model predicting invalid tokens, tracked here as the none and None classes, highlighting the generative collapse completely avoided by our sequential methodology.

These results expose a central trade-off observed in our experiments: the best-performing evaluated sequence is task-dependent. Specifically:

  - Seq1 (MHP-first) produced the highest gains for Mental Health Problem (MHP) detection.
  - Seq2 (Emotion-first) proved superior for Emotion and Sentiment detection.

Table IX: Comprehensive performance comparison across all tasks, contrasting simultaneous Joint-MTL baselines against our proposed Sequential Fine-Tuning methodology.

| | **Joint** | | **Sequential** | | | |
|---|---|---|---|---|---|---|
| **Task** | **Zero-Shot Joint (Mj0)** | **Trained Joint-MTL (Mj)** | **Base Model (M0) F1-Score** | **Single-Task Baseline (M1 / M1e)** | **Best Final Model (Sequence)** | **Improvement** |
| **MHP Detection** | **0.0094** | 0.1953 | 0.6188 | 0.7388 (M1) | 0.7624 ± 0.0405 (Seq1) | **+0.1436** |
| **Emotion Detection** | **0.2545** | 0.5446 | 0.2930 | 0.6004 (M1e) | 0.6052 ± 0.0391 (Seq2) | **+0.3122** |
| **Sentiment Detection** | **0.0118** | 0.7741 | 0.5647 | N/A* | 0.9459 ± 0.0215 (Seq2) | **+0.3812** |

*Note: Sentiment detection was evaluated as a tertiary auxiliary task; therefore, an isolated single-task baseline was not trained for this specific objective.

### *4.5. Limitations and Challenges (RQ4)*

Although the proposed sequential MTL framework demonstrates strong performance, it has several limitations:

1. **Task Interference/Negative Transfer:** The performance degradation observed, most notably for emotion detection in Seq1 and MHP detection in Seq2 after sentiment tuning, is a clear sign of negative transfer or task interference. While PEFT, by freezing the base model, is designed to mitigate severe forgetting, our results show that the LoRA adapters' limited capacity remains susceptible to interference. The adapter weights are overwritten to optimize for the newest task (e.g., simple binary sentiment), potentially harming the more nuanced representations required for previously learned, more complex tasks (e.g., multi-label emotion).
2. **Critical Sensitivity to Task Ordering:** Our findings starkly illustrate that there is no "one-size-fits-all" sequence. The optimal order is entirely dependent on the primary performance objective. This complexity means that deploying such a model requires a deliberate, goal-oriented choice of curriculum, which may not be ideal if balanced performance across all tasks is desired.

3. **Impact of Severe Data Imbalance**: The model's persistent failure to detect several minority emotion classes ('joy', 'calmness', etc.) and poor performance on the 'Positive' sentiment class highlight a fundamental limitation. Even a strong LLM cannot overcome a near-total absence of data for certain labels. This underscores that model performance is still fundamentally constrained by the quality and balance of the training data.

4. **Diminishing Returns on Learning:** In several instances, particularly for sentiment detection in Seq1, performance plateaued very early. This suggests that after an initial knowledge transfer, subsequent direct fine-tuning on a highly related or simpler task may offer no additional benefit, indicating that the given data and model capacity have reached a performance ceiling.

5. **Experimental and Computational Constraints:** Due to the severe computational overhead required to sequentially fine-tune an 8-Billion parameter model, experiments were conducted using single training runs. However, the 95% confidence intervals provide additional estimates of test-set uncertainty around the reported Micro-F1 scores. Since the experiments were conducted using single training runs, these intervals do not account for run-to-run variation caused by stochastic training. Furthermore, the use of sequential, non-stratified data splits and a single dataset limits cross-domain generalization. Finally, deliberately restricting auxiliary tasks to 3 epochs (compared to 10 for the primary task) to prevent overfitting introduces a potential experimental confound, as ordering effects may be partly entangled with the differing epoch budgets.

### *4.6. Qualitative Analysis via Confusion Matrices*

To provide a more granular view of model performance, Figures 9-11 display the normalized confusion matrices for the best-performing final model on each of the three tasks.

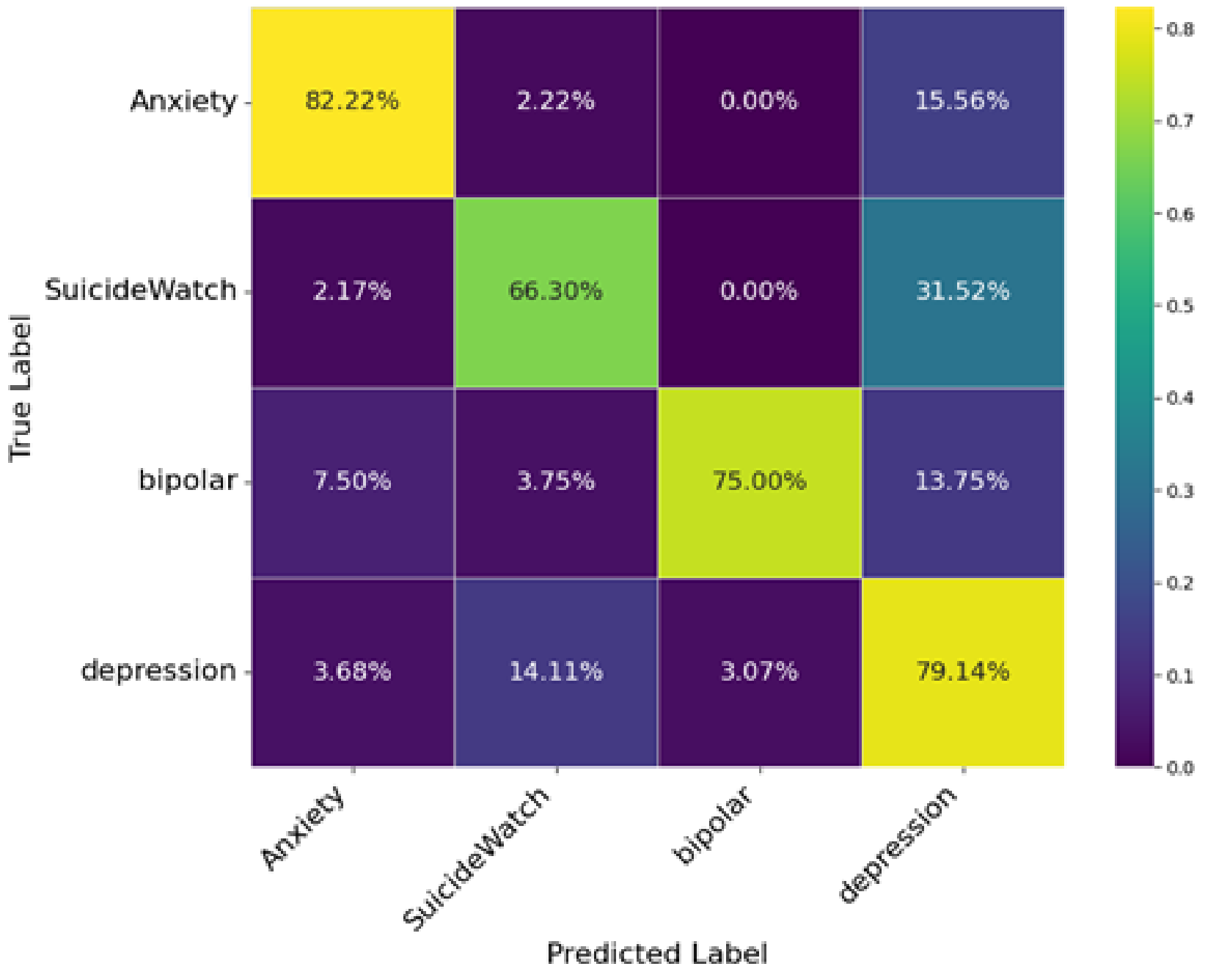

Figure 9: Normalized Confusion Matrix for Mental Health Problem Detection

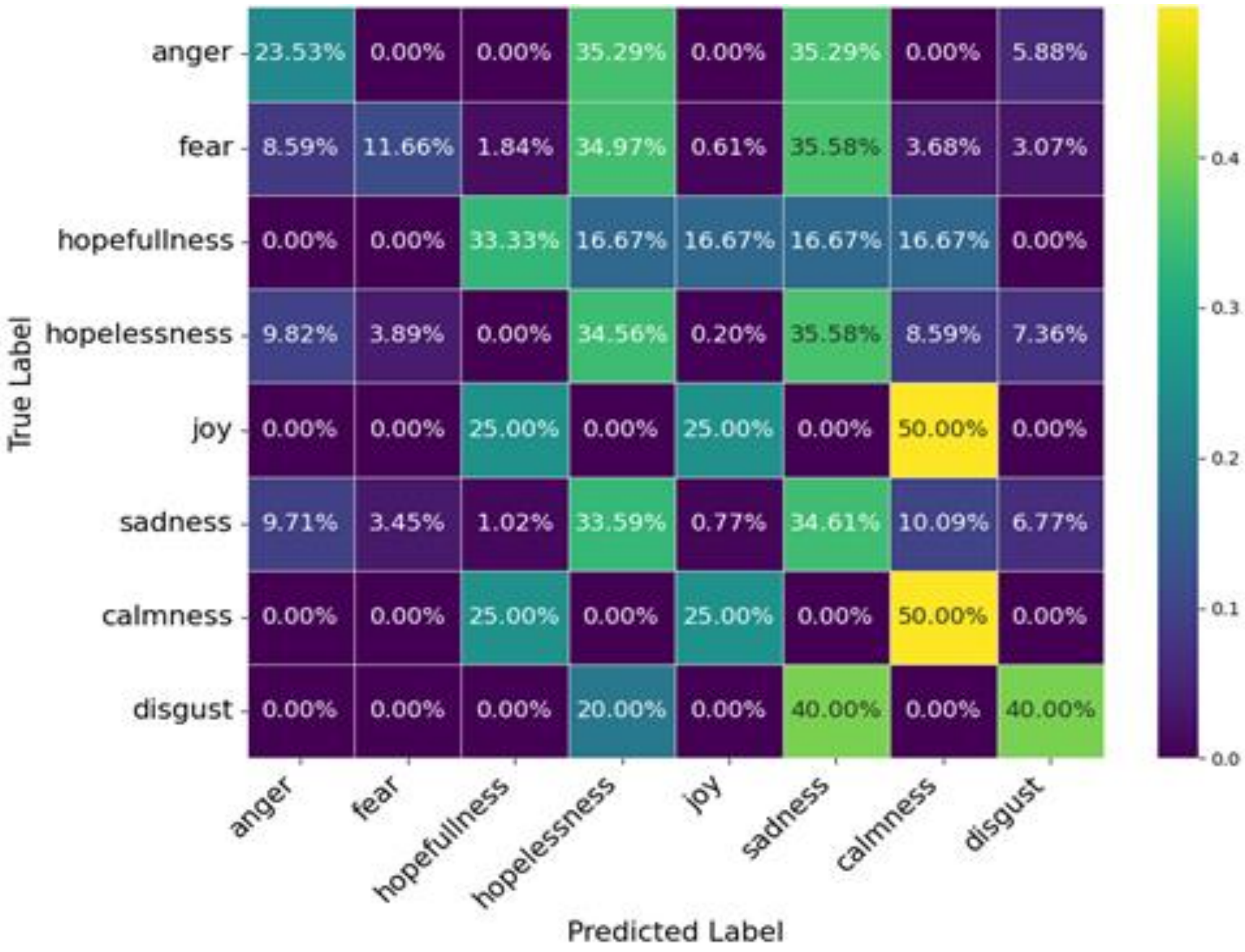

Figure 10: Normalized Confusion Matrix for Emotion Detection

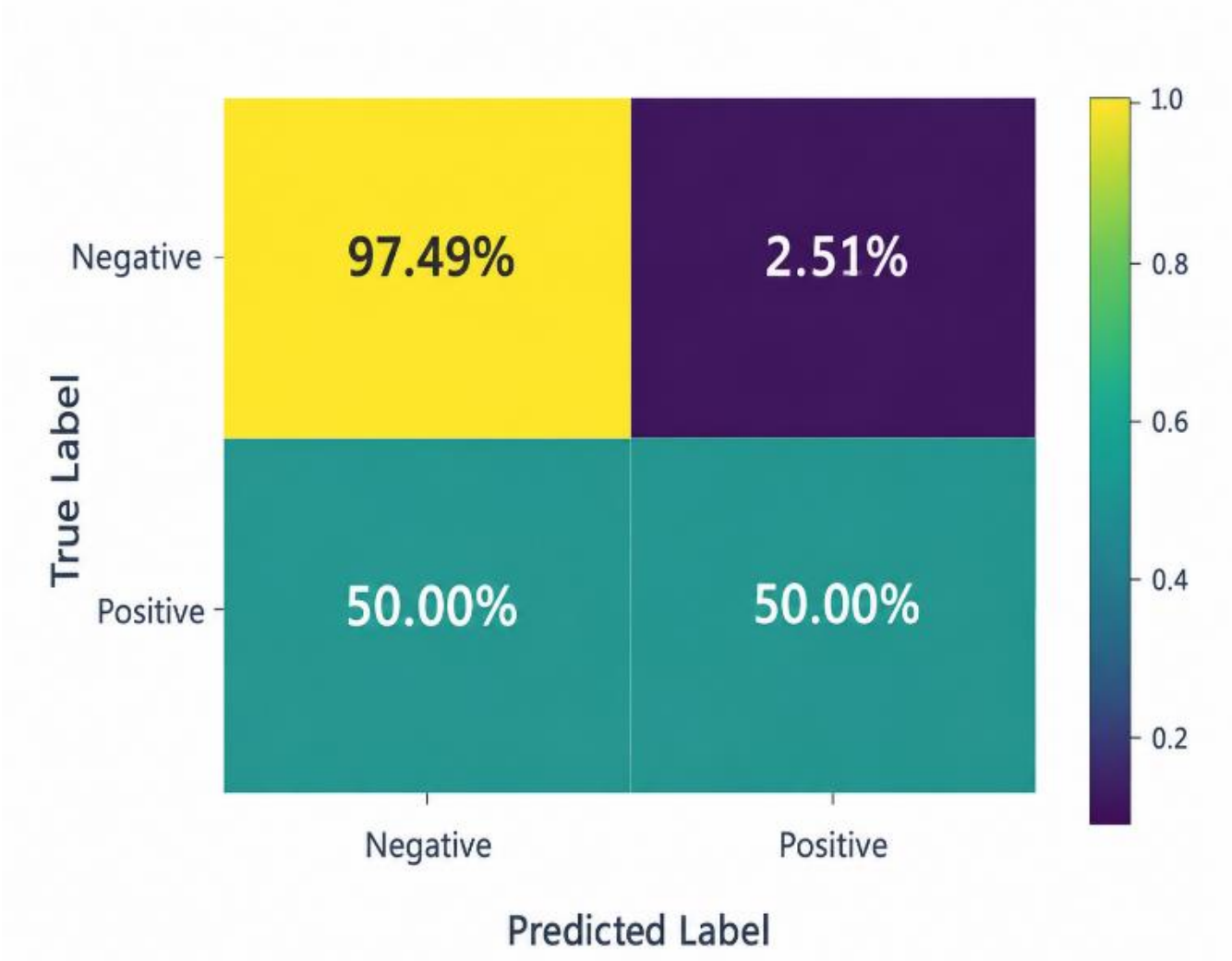

Figure 11: Normalized Confusion Matrix for Sentiment Analysis

These visualizations offer qualitative insights that complement the aggregate F1-scores, revealing the specific nature of the models' successes and failures. For the Mental Health Problem Detection task in Figure 9, the confusion matrix shows strong performance on the diagonal, with high recall for Anxiety (82.22%) and Depression (79.14%). Critically, it also reveals logical, semantically-driven errors; for instance, a significant portion of SuicideWatch posts (31.52%) are misclassified as Depression, which is a clinically related and often co-occurring condition. The confusion matrix in Figure 10 for Emotion Detection underscores the difficulty of this multi-label task. It shows that the model struggles to distinguish between many classes and exhibits a strong bias towards predicting dominant emotions like 'hopelessness' and 'sadness', while rarely predicting minority classes, which visually explains the low overall F1-score for this task. Finally, the Sentiment Analysis confusion matrix in Figure 11 precisely illustrates the impact of severe data imbalance. The model achieves near-perfect recall (97.49%) for the majority 'Negative' class but performs at chance level (50.00%) for the minority 'Positive' class, failing to effectively learn its features. As expected, the vast majority of reviews are labeled as negative because they are related to mental health problems, and consequently, the F1-score of the negative class is higher than the positive class.

To further contrast our proposed methodology against simultaneous multi-tasking, Figures 12–14 display the normalized confusion matrices for the Trained Joint-MTL baseline (Mj). Unlike the structured predictions of our Sequential models, the Joint-MTL matrices visually expose severe task interference and generative collapse. Most notably, the MHP and Sentiment matrices (Figures 12 and 14) clearly reveal predictions falling into the hallucinated "None" column—a failure mode absent in the sequential approach. Furthermore, the Joint-MTL Emotion matrix (Figure 13) demonstrates an extreme collapse toward the majority 'sadness' class, completely ignoring several minority affective states. This qualitative analysis reinforces our numerical findings: sequential task isolation provides the necessary cognitive constraints for reliable generative formatting.

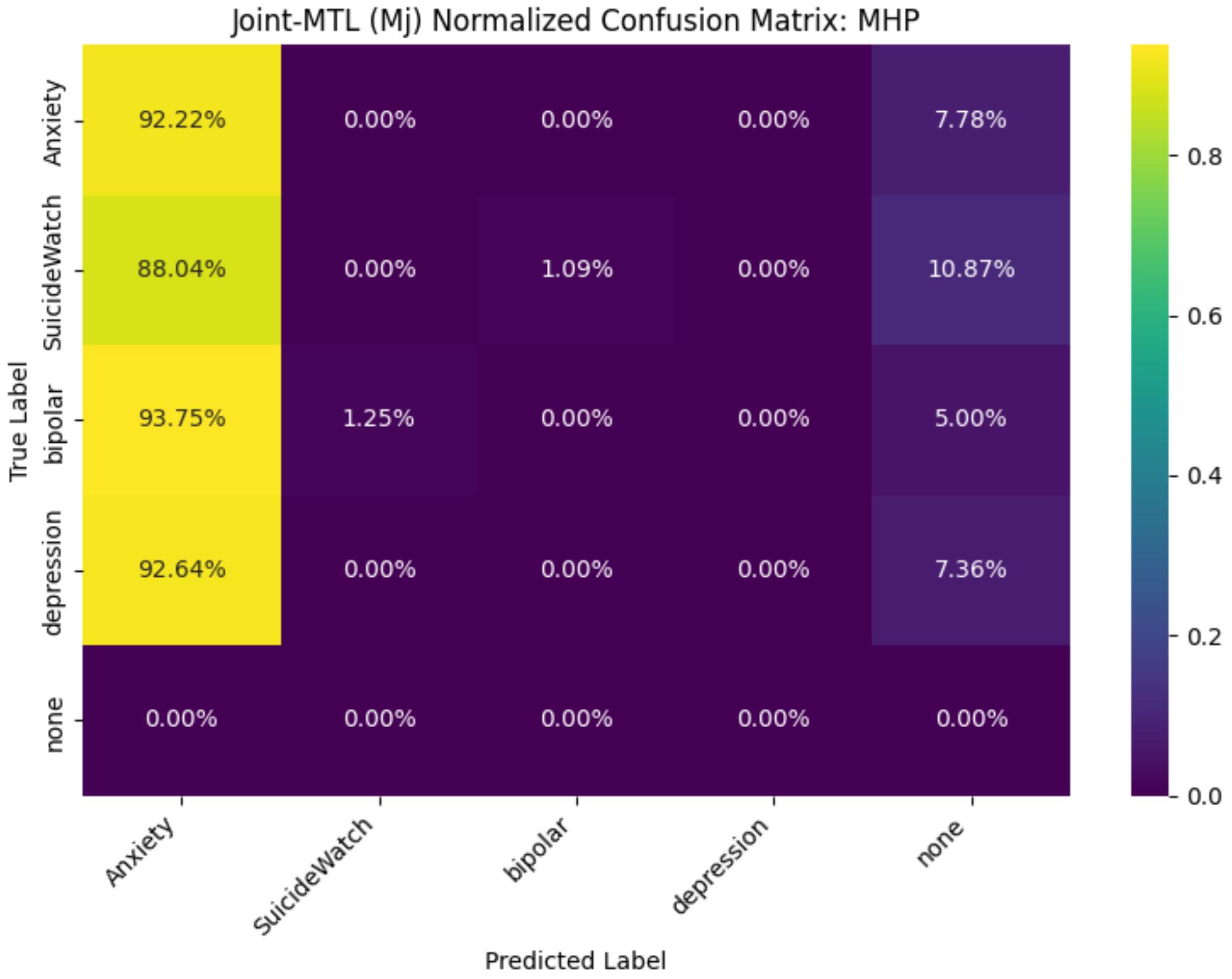

Figure 12: Normalized Confusion Matrix for Mental Health Problem Detection using the Trained Joint-MTL baseline (Mj).

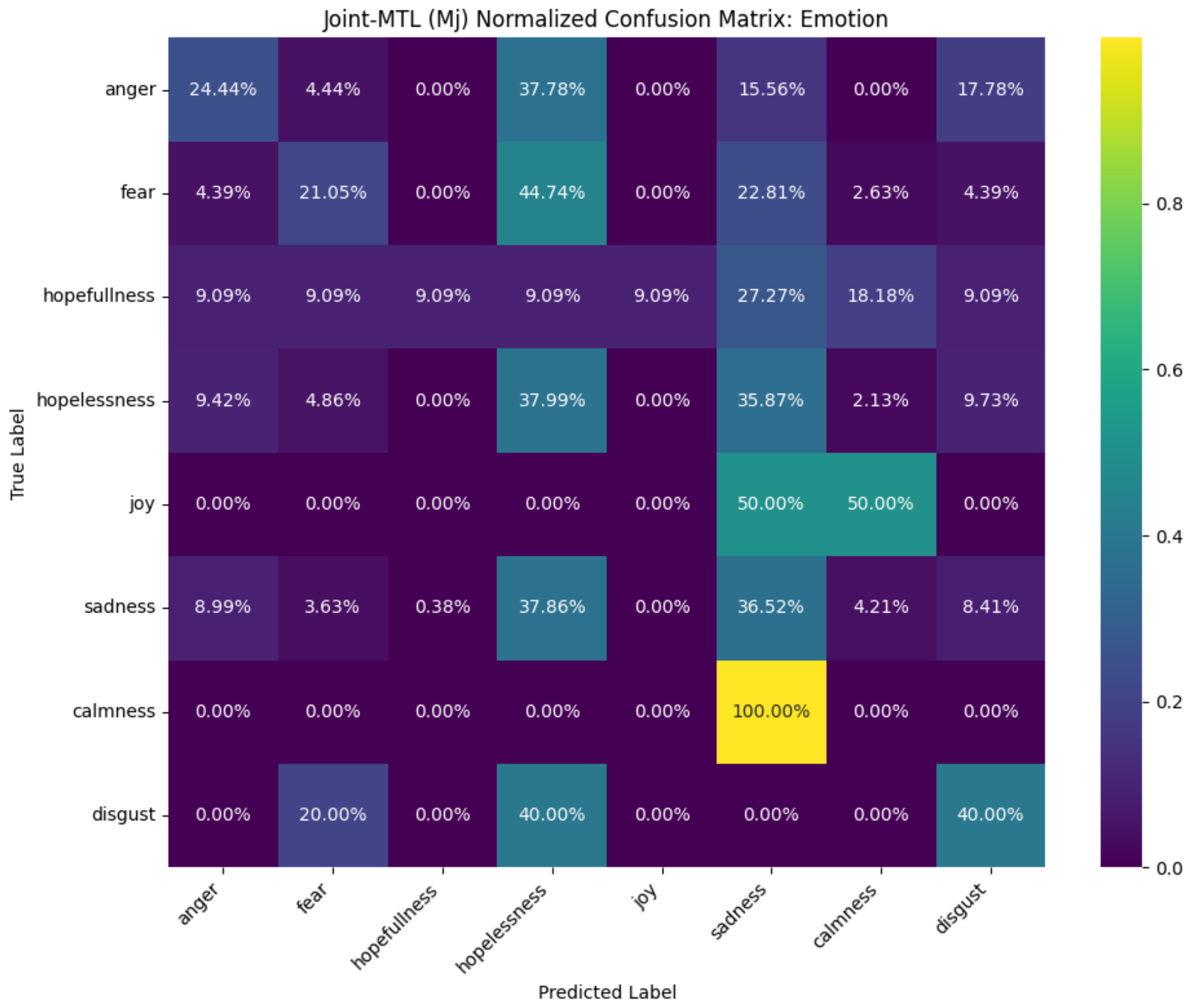

Figure 13: Normalized Confusion Matrix for Emotion Detection using the Trained Joint-MTL baseline (Mj).

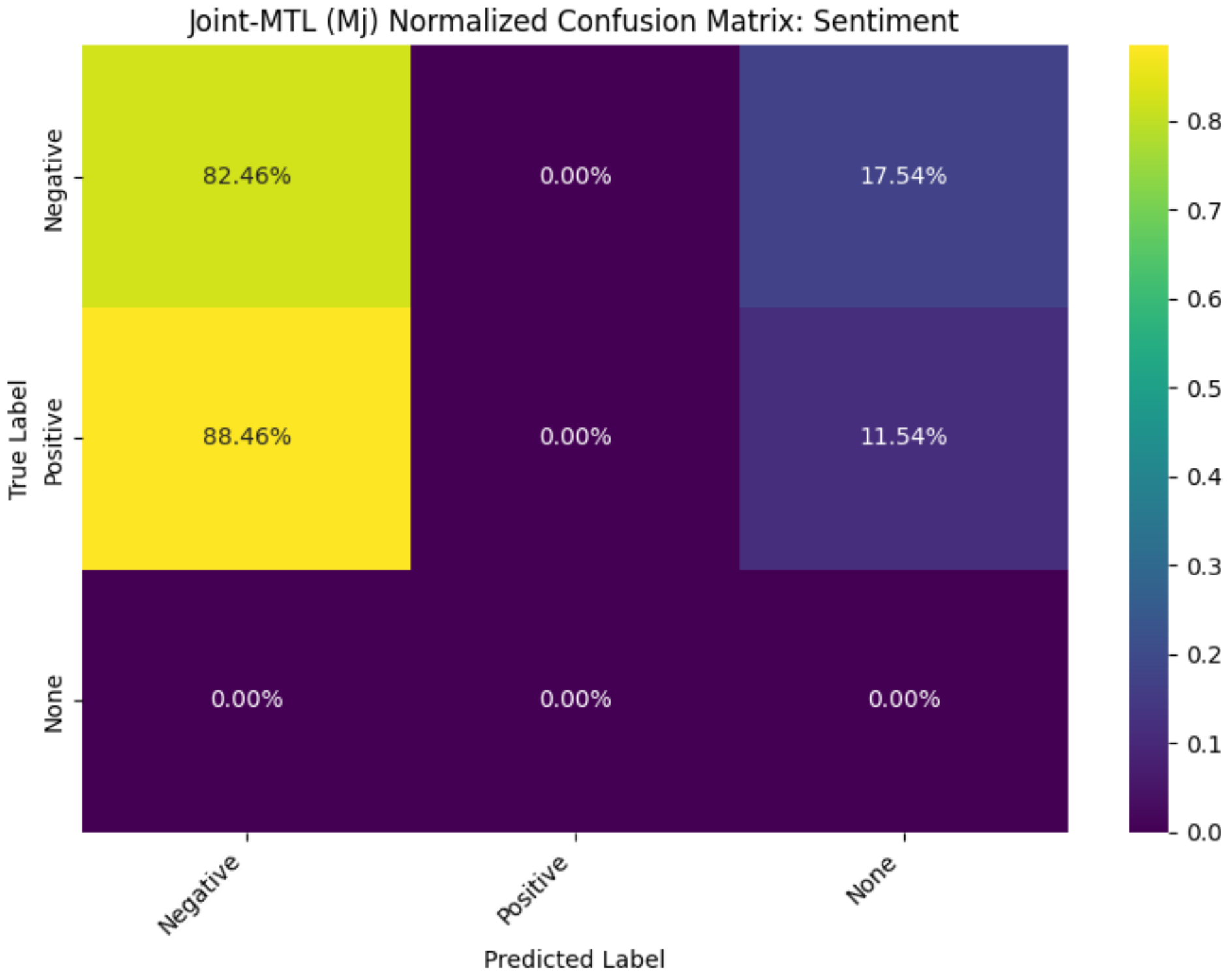

Figure 14: Normalized Confusion Matrix for Sentiment Analysis using the Trained Joint-MTL baseline (Mj).

### *4.7. Impact of Class Imbalance on Minority Metrics:*

Micro-F1 provides an aggregate performance measure but can mask substantial deficits for minority classes in severely imbalanced datasets. Therefore, Macro-F1 and per-class F1-scores are also considered when interpreting the auxiliary-task results. In Emotion Detection, Seq1 achieves a Micro-F1 of 0.4385 but a much lower Macro-F1 of 0.1950, with F1 = 0 for hopefulness, joy, calmness, and disgust. Similarly, Seq2 achieves a Micro-F1 of 0.6052 but a Macro-F1 of only 0.3159, despite a strong F1 of 0.8341 for the dominant Sadness class.

These disparities indicate that the observed Micro-F1 improvements may be influenced primarily by the dominant classes and should therefore not be interpreted as uniform positive transfer across emotion categories. The severe scarcity of minority-class examples is an important confounding factor in the transfer analysis. No class-balancing strategy was applied, as the experiments were designed to evaluate sequential transfer under the original data distribution. Future work should investigate class weighting, resampling, and threshold optimization using more balanced datasets to determine whether minority-class performance and the robustness of the observed transfer effects can be improved.

# 5. Discussion on Results

The results presented in Section 4 demonstrate that sequential MTL using PEFT is a viable and computationally efficient strategy for adapting LLMs such as Llama 3.1 8B for interconnected Mental health problem detection from text. However, the findings also reveal significant complexities regarding knowledge transfer, task interference, and the critical influence of task ordering on performance outcomes. The following subsections discuss these findings in detail.

### *5.1. Efficacy of Sequential PEFT*

Across both experimental sequences, fine-tuning with PEFT consistently and substantially improved Llama 3.1's performance over its zero-shot baseline (M0) on all three tasks. This aligns with the broader PEFT literature [3] [16], underscoring its utility for specializing LLMs with minimal parameter updates. Notably, for Mental Health Problem (MHP) Detection, the Seq1 yielded a final Micro F1-score (0.7624) that surpassed the base LLM. This suggests that for a relatively well-defined primary clinical task, the rich pre-trained representations of an LLM, when efficiently adapted, can deliver superior performance.

### *5.2. Knowledge Transfer Dynamics*

Our investigation into sequential MTL highlighted complex knowledge transfer dynamics:

- Strong Initial Positive Transfer: The experiments revealed substantial positive transfer effects during the early stages of sequential fine-tuning. In both task sequences, learning one mental health-related task improved performance on subsequent auxiliary tasks, indicating that the representations learned from disorder detection, emotion recognition, and sentiment analysis are highly interconnected. These findings suggest that shared psychological and affective patterns across tasks provide an effective inductive bias for sequential multi-task adaptation, consistent with prior studies on the relatedness of mental health and emotional constructs [29] [31].

- Diminishing Returns and Negative Transfer/Interference: Although the primary MHP task in Seq1 continued to benefit from auxiliary-task tuning, the gains for auxiliary tasks were less stable. In particular, Emotion Detection showed only marginal improvement after direct emotion fine-tuning and slightly degraded following sentiment tuning, suggesting task interference between simpler binary sentiment learning and more complex multi-label emotion representations. Similarly, Sentiment Detection plateaued early in Seq1, indicating either a performance ceiling or that the representations learned from MHP detection were already sufficient for this task. These findings suggest task interference and representational overwriting during sequential adaptation, aligning with broader concerns regarding parameter competition in shared adaptation spaces [11] [12], demonstrating that even within a PEFT framework, the limited capacity of LoRA adapters remains susceptible to interference from subsequent tasks.

### *5.3. Task Ordering and Sequential MTL Strategy*

The comparison between the Seq1 and Seq2 sequences demonstrated that task ordering significantly influences both primary and auxiliary task performance in sequential MTL. Starting the task with MHP (Seq1) achieves the highest MHP F1-score (0.7624), while starting it with Emotion detection (Seq2) yields a lower MHP F1 (0.6259). The estimated intervals for the two sequences are well separated (Seq1: [0.7219, 0.8029] vs. Seq2: [0.5799, 0.6719]), indicating a substantial difference in the observed performance. However, because each sequence was evaluated using a single training run, these intervals should not be interpreted as evidence of statistical significance across training runs. This suggests that early parameter specialization is beneficial for optimizing performance on a designated primary task. In contrast, Seq2 yields significantly better performance on Emotion Detection (0.6052 vs. 0.4385) and also improves Sentiment classification (0.9459 vs. 0.8635). These transfer dynamics are heavily informed by the task correlation structure (Figure 6): the robust co-occurrence of depression, hopelessness, and sadness provides a natural semantic bridge when transitioning between clinical and affective states. This pattern suggests that early exposure to emotion-related guidance may foster a more adaptable emotional framework, which, in turn, could enhance subsequent sentiment learning. These findings resonate with curriculum learning principles [8] and studies on task sequencing [32], confirming that the ideal sequence depends on the specific performance objectives for each task and their inter-dependencies.

This indicates that sequential PEFT does not simply accumulate knowledge uniformly across tasks, but instead induces task-dependent specialization and transfer effects. When a single task—such as Mental Health Problem Detection—is the main objective, its appearance as the first task of the sequence may end up with a better result; However, when the objective is to jointly optimize multiple related tasks, task ordering becomes a critical design parameter, and sequential PEFT may require careful curriculum design or be complemented with alternative multi-task strategies.

These sequence-dependent outcomes can be further understood through the label correlations presented in Figure 6. Granular emotion labels such as 'sadness' and 'hopelessness' exhibit strong correlations with clinical classifications like 'depression', as well as with negative sentiment. Starting the sequence with Emotion detection (Seq2) allows the model to initially map these fine-grained affective features, building a rich representational vocabulary that smoothly transfers to the simpler binary Sentiment task. Conversely, starting with Mental Health Problem detection (Seq1) forces the LoRA adapters to immediately optimize for broad, complex clinical buckets.

When subsequently asked to perform multi-label Emotion detection, the model experiences interference, as the adapters must decompose those broad clinical representations back into granular emotional states.

Overall, the obtained results provide practical insights for designing compute-efficient multi-task adaptation pipelines for LLMs. These findings indicate that task-order selection can affect not only predictive performance but also the cumulative computational workload of sequential adaptation. Although PEFT and quantization substantially reduce memory and parameter-update requirements, the underlying 8-billion-parameter model still requires GPU-accelerated computation at every training stage.

### *5.4 Comparative Analysis: Task Interference and Generative Collapse in Joint-MTL*

The comparative analysis detailed in **Table IX** starkly illustrates the vulnerabilities inherent to simultaneous multi-tasking in generative LLMs.

- **Zero-Shot Format Collapse:** When evaluated in a Zero-Shot setting, the base Llama-3.1-8B-Instruct model suffered format collapse. Unable to reliably follow multi-step formatting instructions, it achieved near-zero Micro-F1 scores for MHP (0.0094) and Sentiment (0.0118). This confirms that native instruction-tuned models cannot reliably extract and format complex, overlapping psychiatric features without domain-specific fine-tuning.
- **Task Interference and Negative Transfer:** While fine-tuning on the Joint-MTL formulation significantly improved format compliance, the model suffered from task interference. As shown in the detailed classification reports (**Table VIII**), the Joint-MTL model allocated nearly all representational capacity to a single MHP class (Anxiety, achieving 0.9222 recall) while completely failing on bipolar, depression, and SuicideWatch (all scoring 0.0000 F1). This pattern is consistent with negative transfer: the simultaneous gradient signals from Emotion and Sentiment optimization overwhelmed the model's ability to learn the full MHP classification space, collapsing it to a Micro-F1 of just 0.1953. The Sentiment task, being the simplest binary classification, survived Joint-MTL with the least damage (0.7741), yet the model completely failed on the minority Positive class (0.0000 recall), revealing a severe majority-class bias induced by multi-task competition.
- **Format Compliance and Generative Collapse:** A qualitatively distinct failure mode emerged in the Joint-MTL evaluation. Because Llama 3.1 is an autoregressive generative model, it is susceptible to hallucinating out-of-distribution tokens when cognitively overloaded. As documented in **Table VIII**, the Joint-MTL model frequently experienced *generative collapse*, hallucinating invalid categorical labels or omitting labels entirely. These invalid outputs are captured as the "None" class (support = 0) in our transparent evaluation reports. In stark contrast, our proposed Sequential Fine-Tuning methodology—by isolating the cognitive load to a single task per fine-tuning stage—achieved 100% format compliance across all evaluation instances, producing zero out-of-vocabulary predictions. This finding has critical implications for clinical deployment, where hallucinated diagnoses could lead to patient harm.

These results empirically validate our core thesis: the evaluated simultaneous multi-task configuration exhibited substantial task interference for the model's representational capacity, with simpler tasks cannibalizing the gradient signal needed by more complex tasks. The evaluated sequential task-isolation configuration substantially reduced accuracy degradation and invalid outputs compared with the evaluated Joint-MTL configuration.

# 6. Ethical Considerations

Given the highly sensitive nature of mental health text analysis, it is imperative to address the ethical limitations and potential risks associated with deploying LLMs in this domain. First, **privacy and anonymity** are paramount. While the SWMH dataset utilizes publicly available, anonymized social media posts, deploying such models in real-world applications requires stringent data governance to prevent the deanonymization of vulnerable individuals. Second, we emphasize the risks of misclassification harms. As demonstrated by our Macro-F1 analysis, the model exhibits bias toward majority classes and struggles with underrepresented emotional states. In a real-world setting, a false negative (failing to detect severe distress such as SuicideWatch) or a false positive could lead to inappropriate interventions or missed support opportunities. Finally, we explicitly state that this sequentially adapted LLM is an analytical research tool, not a clinical diagnostic system. Predictions generated by NLP models must never replace professional psychiatric evaluation. Any future deployment of such technologies must enforce strict "human-in-the-loop" oversight, ensuring that algorithmic outputs serve only to flag potential distress for review by qualified mental health professionals.

# 7. Conclusion and Future Work

This paper investigated the adaptation of the Llama 3.1 8B Instruct model for interconnected mental health text analysis tasks, Mental Health Problem Detection, Multi-Label Emotion Detection, and Sentiment detection, using a sequential multi-task learning (MTL) strategy with Parameter-Efficient Fine-Tuning (PEFT). Our findings demonstrate that sequential PEFT significantly enhances LLMs' capabilities for these specialized tasks, surpassing their zero-shot performance and achieving competitive results on a primary clinical task.

The study revealed complex knowledge transfer dynamics: strong positive transfer was observed from initial fine-tuning on mental health problem labels to auxiliary emotion and sentiment tasks. However, subsequent stages showed diminishing returns for the primary task and, in some cases, notably emotion detection in the MHP-first sequence, potential negative transfer or task interference. Crucially, our comparative analysis of two distinct fine-tuning sequences (MHP→Emo→Sent vs. Emo→MHP→Sent) highlighted that the optimal sequential order is highly task-dependent. While prioritizing mental health problem detection first yielded the best results for that task, an Emotion-first sequence proved superior for both Emotion and Sentiment detection. This underscores that there is no single optimal sequence for all tasks in sequential MTL, and that the choice of curriculum significantly affects performance trade-offs. The persistent difficulty in detecting severely imbalanced emotion and sentiment classes also emphasized the critical role of data characteristics.

From an HPC perspective, the current study demonstrates that efficient adaptation of large foundation models is closely tied to hardware-aware optimization. The combination of QLoRA, 4-bit quantization, mixed-precision computation, gradient checkpointing, and sequential task adaptation reduces the resource requirements of LLM adaptation while preserving competitive predictive performance.

In a nutshell, sequential MTL with PEFT provides a powerful, computationally efficient approach for specializing LLMs in the mental health domain. However, its successful application requires careful consideration of task-relatedness, strategic task ordering based on primary objectives, and awareness of potential interference and data limitations. Beyond improvements in task performance, the findings demonstrate that sequential PEFT-based adaptation provides a practical balance between computational efficiency and task specialization for LLM models. This is particularly relevant for resource-aware AI systems, where minimizing GPU memory usage and training overhead is essential for scalable deployment. Crucially, our comprehensive baseline comparisons conclusively demonstrate that while simultaneous Joint Multi-Task Learning induces severe negative transfer and generative format collapse—hallucinating invalid diagnostic labels—our proposed sequential methodology eliminates these failure modes. By sequentially isolating feature extraction, the model achieves state-of-the-art accuracy with perfect format compliance, suggesting that sequential task isolation may be advantageous for reliable generative mental-health assessment under the evaluated experimental setting. Building on the insights and limitations identified in this research, several avenues for future work emerge:

- Systematic Exploration of Task Orderings and Curriculum Design: Investigating a broader range of task permutations and principled curriculum learning strategies (e.g., ordering by task complexity or data availability) to identify optimal paths for knowledge accumulation and minimizing negative transfer in sequential MTL for LLMs, as well as validating these architectures across additional multi-annotated clinical datasets to ensure broad generalizability.
- Advanced Techniques for Mitigating Negative Transfer and Catastrophic Forgetting: Exploration and adaptation of continual learning techniques (e.g., Elastic Weight Consolidation, replay mechanisms tailored for PEFT) to specifically address and alleviate task interference and catastrophic forgetting observed in sequential LLM fine-tuning.
- Granular Analysis of PEFT Adapter Dynamics: Investigating how LoRA adapter parameters evolve and specialize during sequential MTL. Analyzing parameter overlaps or divergence across stages could provide deeper insights into how knowledge is transferred or interfered with at the model parameter level.
- Exploring Advanced Class Imbalance Mitigation in LLMs: To address the severe performance deficits observed on minority affective classes, future research should integrate class-balancing strategies directly into the LLM fine-tuning architecture. This includes investigating the efficacy of specialized loss functions (e.g., Focal Loss or class-weighted cross-entropy) within a PEFT framework, as well as exploring dynamic oversampling techniques for instruction-tuning datasets to prevent majority-class dominance.

# Declaration of Generative AI and AI-assisted Technologies in the Writing Process

During the preparation of this work, the authors used ChatGPT to improve the manuscript's readability and language. After using this tool, the authors reviewed and edited the content as needed and take full responsibility for the published article.

# Declaration of competing interest

The authors declare that they have no known competing financial interests or personal relationships that could have influenced the work reported in this paper.

# Data availability

The datasets generated and/or analyzed during the current study are available from the first author on reasonable request.

# Funding

Not applicable.

# Author Contributions

First author: Conceptualization, Software development, Validation, Writing Original manuscript.
Second author: Advisorship, writing the original manuscript, reviewing and editing the paper.
Third author: Co-Advisorship, writing the original manuscript, reviewing and editing the paper.

# References

[1] World Health Organization, "Mental disorders," WHO Fact Sheet, 2025. [Online]. Available: https://www.who.int/news-room/fact-sheets/detail/mental-disorders.

[2] Llama Team, AI @ Meta. (2024). The Llama 3 Herd of Models. arXiv preprint arXiv:2407.21783v3 [cs.AI].

[3] Hu, E. J., Shen, Y., Wallis, P., Allen-Zhu, Z., Li, Y., Wang, S., Wang, L., & Chen, W. (2022). LoRA: Low-rank adaptation of large language models. Proceedings of the International Conference on Learning Representations.

[4] Armah, C., & Dehkharghani, R. (2024). Multi-task learning on mental disorder detection, sentiment analysis, and emotion detection using social media posts. 2024 8th International Symposium on Innovative Approaches in Smart Technologies (ISAS), 1–10. https://doi.org/10.1109/ISAS64331.2024.10845733

[5] Caruana, R. (1997). Multitask learning. Machine Learning, 28(1), 41–75.

[6] Ruder, S. (2017). An overview of multi-task learning in deep neural networks. arXiv preprint arXiv:1706.05098.

[7] Zhang, Y., & Yang, Q. (2021). A survey on multi-task learning. IEEE Transactions on Knowledge and Data Engineering, 34(12), 5586–5609.

[8] Bengio, Y., Louradour, J., Collobert, R., & Weston, J. (2009). Curriculum learning. In Proceedings of the 26th Annual International Conference on Machine Learning (pp. 41–48).

[9] Ilias, L., & Askounis, D. (2023). Multitask learning for recognizing stress and depression in social media. Online Social Networks and Media, 37, 100270.

[10] McCloskey, M., & Cohen, N. J. (1989). Catastrophic interference in connectionist networks: The sequential learning problem. In Psychology of Learning and Motivation (Vol. 24, pp. 109–165). Academic Press.

[11] French, R. (1999). Catastrophic forgetting in connectionist networks. Trends in Cognitive Sciences, 3(4), 128–135. https://doi.org/10.1016/S1364-6613(99)01294-2

[12] Ke, Z., Huang, H., Chen, X., & Liu, B. (2021). Continual learning for aspect-sentiment classification. Proceedings of the 2021 Conference on Empirical Methods in Natural Language Processing, 6871–6883.

[13] Haque, N. (2025). Catastrophic forgetting in LLMs: A comparative analysis across language tasks. arXiv preprint arXiv:2504.01241.

[14] Song, S. (2025). How to alleviate catastrophic forgetting in LLMs finetuning? Hierarchical layer-wise and element-wise regularization. arXiv. https://doi.org/10.48550/arXiv.2501.13669

[15] X. Zhang, K. Wang, T. Hu and H. Ma, "Efficient Knowledge Transfer in Multi-Task Learning through Task-Adaptive Low-Rank Representation," 2025 IEEE International Conference on Multimedia and Expo (ICME), Nantes, France, 2025, pp. 1-6, doi: 10.1109/ICME59968.2025.11210217.

[16] Dettmers, T., Pagnoni, A., Holtzman, A., & Zettlemoyer, L. (2023). QLoRA: Efficient finetuning of quantized LLMs. Advances in Neural Information Processing Systems, 36, 10088–10115.

[17] Wang, L., Chen, S., Jiang, L. et al. (2025). Parameter-efficient fine-tuning in large language models: A survey of methodologies. Artificial Intelligence Review, 58, 227. https://doi.org/10.1007/s10462-025-11236-4

[18] Houlsby, N., Giurgiu, A., Jastrzebski, S., Morrone, B., De Laroussilhe, Q., Gesmundo, A., ... & Gelly, S. (2019). Parameter-efficient transfer learning for NLP. In International Conference on Machine Learning (pp. 2790–2799). PMLR.

[19] Li, X. L., & Liang, P. (2021, August). Prefix-tuning: Optimizing continuous prompts for generation. In Proceedings of the 59th annual meeting of the association for computational linguistics and the 11th international joint conference on natural language processing (volume 1: Long papers) (pp. 4582-4597).

[20] Lester, B., Al-Rfou, R., & Constant, N. (2021). The power of scale for parameter-efficient prompt tuning. In Proceedings of the 2021 Conference on Empirical Methods in Natural Language Processing (pp. 3045–3059). Association for Computational Linguistics.

[21] De Choudhury, M., Gamon, M., Counts, S., & Horvitz, E. (2013). Predicting depression via social media. Proceedings of the 7th International AAAI Conference on Weblogs and Social Media, 128–137.

[22] Coppersmith, G., Dredze, M., Harman, C., Hollingshead, K., & Mitchell, M. (2015). CLPsych 2015 shared task: Depression and PTSD on Twitter. Proceedings of the 2nd Workshop on Computational Linguistics and Clinical Psychology, 31–39.

[23] Calvo, R. A., Milne, D. N., Hussain, M. S., & Christensen, H. (2017). Natural language processing in mental health applications using non-clinical texts. Natural Language Engineering, 23(5), 649–685. https://doi.org/10.1017/S1351324916000383

[24] Kermani, A., Perez-Rosas, V., & Metsis, V. (2025, May). A systematic evaluation of llm strategies for mental health text analysis: Fine-tuning vs. prompt engineering vs. rag. In Proceedings of the 10th Workshop on Computational Linguistics and Clinical Psychology (CLPsych 2025) (pp. 172-180).

[25] Ji, S., Li, X., Huang, Z. et al. Suicidal ideation and mental disorder detection with attentive relation networks. Neural Comput & Applic (2021). https://doi.org/10.1007/s00521-021-06208-y

[26] Dehkharghani, R., Armah, C., Ozerceli, O., & Treter, A. (2023). Extended SWMH dataset. https://github.com/flightmaestro/extendedswmh

[27] Ji, S., Zhang, T., Ansari, L., Fu, J., Tiwari, P., & Cambria, E. (2022). MentalBERT: Publicly available pretrained language models for mental healthcare. Proceedings of the Language Resources and Evaluation Conference, 7184–7190.

[28] Yang, K., Zhang, T., Kuang, Z., Xie, Q., Huang, J., & Ananiadou, S. (2024). MentaLLaMA: Interpretable mental health analysis on social media with large language models. Proceedings of the ACM Web Conference 2024, 4489–4500.

[29] Pathak, A., Bhattacharjee, S., Saha, T., & Saha, S. (2025). Do sentiment and emotion affect mental health? A multi-task classification framework for comprehensive understanding of mental health, emotion, and sentiment from motivational conversations. ACM Transactions on Computing for Healthcare, 6(2), 1-16. https://doi.org/10.1145/3704740

[30] Ali, M., Lucasius, C., Patel, T. P., Aitken, M., Vorstman, J., Szatmari, P., ... & Kundur, D. (2025, December). Speech as a multimodal digital phenotype for multi-task llm-based mental health prediction. In 2025 International Conference on Smart Multimedia (ICSM) (pp. 1-6). IEEE.

[31] Akhtar, M. S., Chauhan, D., Ghosal, D., Poria, S., Ekbal, A., & Bhattacharyya, P. (2019). Multi-task learning for multi-modal emotion recognition and sentiment analysis. Proceedings of the 2019 Conference of the North American Chapter of the Association for Computational Linguistics: Human Language Technologies, 370–379.

[32] Liu, Y., Zhang, X., Huang, S., & Yang, D. (2024). Sequential multi-modal learning for emotion understanding in conversational settings. Proceedings of the 2024 Conference on Empirical Methods in Natural Language Processing, 2156–2167.

[33] The Llama 3.1 8B Instruct Model. https://huggingface.co/meta-llama/Llama-3.1-8B-Instruct

[34] Z. Tang, H. Yang and C. Y. -C. Chen, "Weakly Supervised Posture Mining for Fine-Grained Classification," 2023 IEEE/CVF Conference on Computer Vision and Pattern Recognition (CVPR), Vancouver, BC, Canada, 2023, pp. 23735-23744, doi: 10.1109/CVPR52729.2023.02273.

[35] J. Huang et al., "ConsistentID: Portrait Generation With Multimodal Fine-Grained Identity Preserving," in IEEE Transactions on Pattern Analysis and Machine Intelligence, vol. 48, no. 5, pp. 5639-5654, May 2026, doi: 10.1109/TPAMI.2026.3652557.

[36] Jiehui Huang, Jun Zhou, Zhenchao Tang, Jiaying Lin, Calvin Yu-Chian Chen, TMBL: Transformer-based multimodal binding learning model for multimodal sentiment analysis, Knowledge-Based Systems, Volume 285, 2024, 111346, ISSN 0950-7051, https://doi.org/10.1016/j.knosys.2023.111346. (https://www.sciencedirect.com/science/article/pii/S0950705123010948)

[37] Zhenchao Tang, Fang Wang, Haohuai He, Jiale Zhou, Tianxu Lv, Jun Zhu, Shouzhi Chen, Minghao Yang, Yu Wang, Jiayang Wu, Yidong Song, Yaokun Li, Jiehui Huang, Bing He, Jianhua Yao: Aligning LLMs with Biomedical Knowledge using Balanced Fine-Tuning https://doi.org/10.48550/arXiv.2511.21075